%% file: neurips_2026.tex
\documentclass{article}

\usepackage[main, final]{neurips_2026}

\usepackage[utf8]{inputenc} 
\usepackage[T1]{fontenc}    
\usepackage{hyperref}       
\usepackage{url}            
\usepackage{booktabs}       
\usepackage{amsfonts}       
\usepackage{nicefrac}       
\usepackage{microtype}      
\usepackage{xcolor}         

\title{DRTriton: Large-Scale Synthetic Data Driven Reinforcement Learning for Triton Kernel Generation}

\author{%
  Siqi Guo \\
  Texas A\&M University\\
  \texttt{siqi@tamu.edu} \\
  \And
  Ming Lin \\
  Oracle \\
  \texttt{linming04@gmail.com} \\
  \And
  Tianbao Yang\thanks{Corresponding author.} \\
  Texas A\&M University\\
  \texttt{tianbao-yang@tamu.edu} \\
}

\usepackage{graphicx}
\usepackage{makecell}
\usepackage{algorithm,algorithmic}
\usepackage{paralist,amsmath, amssymb,bm}
\usepackage{multirow}

\usepackage{textcase}
\usepackage{booktabs}
\usepackage{fancybox}
\usepackage{mathtools}
\usepackage{xcolor,colortbl}
\usepackage{subcaption}
\usepackage{longtable}
\usepackage{tcolorbox}
\usepackage{listings}
\usepackage{wrapfig}
\usepackage{enumitem}

\definecolor{codegreen}{rgb}{0,0.6,0}
\definecolor{codegray}{rgb}{0.5,0.5,0.5}
\definecolor{codepurple}{rgb}{0.58,0,0.82}
\definecolor{backcolour}{rgb}{0.97,0.97,0.97}
\definecolor{stringcolor}{rgb}{0.75,0.1,0.1}
\definecolor{keywordcolor}{rgb}{0.0,0.0,0.8}

\lstdefinestyle{minimalstyle}{
    backgroundcolor=\color{backcolour},   
    commentstyle=\color{codegreen}\itshape,
    keywordstyle=\color{keywordcolor}\bfseries,
    stringstyle=\color{stringcolor},
    basicstyle=\ttfamily\small,
    breaklines=true,                 
    keepspaces=true,                 
    showspaces=false,                
    showstringspaces=false,
    showtabs=false,                  
    tabsize=2,
    frame=tb,  
    aboveskip=8pt,
    belowskip=8pt,
    xleftmargin=5pt,
    xrightmargin=5pt,
}
\def \S {\mathbf{S}}

\def \x {\mathbf{x}}

\def \x {\mathbf{x}}

\def \1 {\mathbf{1}}

\def \y {\mathbf{y}}

\def \y {\mathbf{y}}

\def \x {\mathbf{x}}

\def \S {\mathcal{S}}

\begin{document}

\maketitle

\begin{abstract}
Developing efficient CUDA kernels is a fundamental yet challenging task in the generative AI industry. Recent research leverages Large Language Models (LLMs) to automatically convert PyTorch reference implementations to CUDA kernels, significantly reducing engineering effort.
State-of-the-art LLMs, such as GPT-5.2 and Claude-Sonnet-4.5, still struggle with this task.
To address this challenge, we propose \textbf{DRTriton}, a scalable learning framework for training LLMs to convert PyTorch programs into highly optimized Triton kernels, which are then compiled to CUDA kernels at runtime. DRTriton consists of three key components: (i) a data synthetic algorithm CSP-DAG that guarantees full coverage and unbiased uniform sampling over the operator space with controlled difficulty; (ii) a curriculum RL framework with decoupled rewards that jointly optimizes conversion success rate and execution speed;  and (iii) a test-time search algorithm that further improves the execution speed of the generated Triton kernels. With a warmup stage of SFT on limited PyTorch-Triton pairs curated using existing LLMs, DRTriton trained by RL on synthesized PyTorch programs generalizes effectively to real-world CUDA kernels that are challenging even for human experts.
Experimental results show that DRTriton-7B achieves speedup over PyTorch on 92\% of KernelBench Level 2 tasks, compared to 23\% for GPT-5.2 and 19\% for Claude-Sonnet-4.5.
\end{abstract}

\section{Introduction}


The rapid growth of generative AI has driven the industry to seek highly optimized CUDA kernels to reduce inference costs. However, developing efficient CUDA kernels remains a significant challenge even for human experts. For instance, the widely used FlashAttention~\cite{dao2022flashattention} library requires years of development effort. As new algorithms and models continue to emerge, it is increasingly impractical to manually create optimized CUDA kernels for each one. Automating efficient CUDA kernel development has become a critical need.

To simplify CUDA kernel development, practitioners often turn to more user-friendly domain-specific languages (DSLs) such as \textit{Triton}~\citep{tillet2019triton}. \textit{Triton} offers an alternative API to CUDA with a PyTorch-like interface, achieving a good balance between development and runtime efficiency. Despite this advantage, writing efficient Triton kernels still requires substantial expertise in GPU programming and involves extensive trial and error. An automated tool capable of converting PyTorch reference implementations into optimized Triton kernels would therefore greatly alleviate this burden.


Thanks to the remarkable code generation capabilities of recent large language models (LLMs)~\citep{chen2021evaluating,roziere2023code,huang2025opencoder}, it is promising to leverage LLMs for this conversion. While being powerful in various aspects, modern open-source and commercial LLMs still fall short in translating PyTorch implementations into efficient Triton kernels and/or CUDA kernels, even on relatively simple tasks~\citep{ouyang2025kernelbench,li2025tritonbench}.


In this work, we focus on converting PyTorch programs to Triton kernels using generative AI technologies. Specifically, we aim to train an LLM for this task. Several recent works~\citep{li2025autotriton,woo2025tritonrl} try to address this challenge by learning on PyTorch programs and their corresponding Triton kernels collected from public code repositories. While showing promising improvements over baseline models, the capacities of these approaches are inherently limited by the following factors:

\textbf{Limited scale of training data.} For example, \citet{li2025autotriton} use 14k samples and \citet{woo2025tritonrl} use 11k samples, which is insufficient for learning complex Triton or CUDA kernels.
    

\textbf{Uncontrolled complexity and quality.} When training with sparse rewards, it is crucial to carefully control the order and difficulty of samples presented to the model. If challenging samples are introduced too early, the model may consistently receive zero rewards, leading to poor gradient signals, wasted training tokens, and potentially misleading updates.



To address these challenges, we introduce a novel approach that leverages synthetic data. Instead of relying on limited real-world PyTorch code, we systematically generate PyTorch programs with varying difficulty and employ curriculum reinforcement learning, allowing the model to progress from simple to complex tasks. Our approach begins by collecting a representative subset of 61 widely used PyTorch operators, including those used in the KernelBench benchmark~\citep{ouyang2025kernelbench}. We then formulate PyTorch program generation as a constraint satisfaction problem over directed acyclic graphs (CSP-DAG), thereby enabling efficient and uniform sampling of valid programs composed of these operators. Based on this synthetic data generation strategy, we present \textbf{DRTriton}, a scalable framework that trains PyTorch-to-Triton LLMs based on large-scale synthetic PyTorch programs. DRTriton consists of three key components, which constitute our main contributions:
\vspace*{-0.1in}
\begin{itemize}[leftmargin=*]
\item {\bf CSP-DAG sampling algorithm for PyTorch code generation.} We propose an efficient algorithm for synthesizing \textbf{valid} PyTorch programs. We formulate the PyTorch program generation as a constraint satisfaction problem (CSP) on directed acyclic graphs (DAGs), which can be solved using CP-SAT solvers~\citep{cpsatlp}. This approach allows us to sample valid PyTorch programs in our code space.
\item \textbf{Large-scale curriculum reinforcement learning with decoupled rewards.} We introduce a curriculum learning scheme and decouple rewards into the conventional reinforcement learning framework. It stabilizes our learning process under sparse reward and improves our learning efficiency, especially in the early stage.
\item \textbf{Test-time search for compositional kernels.} We introduce an effective test-time search strategy that decomposes complex PyTorch programs into several smaller kernels and searches for optimal compositions of kernels. This test-time search strategy substantially improves both conversion success rate and execution speed when the PyTorch program cannot be translated to a single kernel.
\end{itemize}

Leveraging the above techniques, we learn a DRTriton-7B model initialized from Qwen-2.5-Coder-7B-Instruct, using 2,026 single-operator PyTorch-Triton pairs curated by DeepSeek-R1 and GPT-5.2, followed by curriculum RL on 100,000 synthetic PyTorch programs. The resulting model demonstrates strong performance: (i) On our synthetic benchmark with 20 operators, it achieves 99\% accuracy, with 86\% of generated Triton kernels outperforming the original PyTorch implementations. (ii) On KernelBench Level 3, the model attains 76\% accuracy, with 54\% of generated Triton kernels faster than the original PyTorch code and 34\% faster than kernels produced by \texttt{torch.compile}.


\section{Related Work}
\label{sec:related}

\textbf{Kernel Compilers.\quad}
Conventional kernel optimization relies on human-crafted optimization algorithms. TVM~\citep{chen2018tvm} and Ansor~\citep{zheng2020ansor} utilize machine learning to search for optimized schedules with manual specified patterns. 
TorchInductor~\citep{ansel2024pytorch} has emerged as the default compiler (used by \texttt{torch.compile}) since PyTorch 2.0. It maps PyTorch models to Triton kernels through pattern matching and loop fusion. While effective, these compilers are bounded by predefined heuristics and templates. 
Our approach learns from uniformly sampled PyTorch code, avoiding reliance on pre-defined heuristics.

\textbf{GPU Kernel Generation with LLMs.\quad}
To overcome the rigidity of conventional compilers, recent works explored LLMs for direct kernel generation. \citet{fischeskernelllm} curated a dataset of 25k paired examples of PyTorch modules with their equivalent Triton kernel implementations, and then trained a model named KernelLLM by SFT. However, their model suffers from hallucination and ``fake" kernels. To improve KernelLLM, two recent works leveraged RL to fine-tune LLMs based on hand-curated data~\citep{li2025autotriton,woo2025tritonrl}. These approaches require curation of training samples from public datasets, which limits their scalability.

\textbf{Multi-agent systems.\quad}
Multi-agent systems have been proposed to coordinate planning, generation and verification~\citep{hong2025autocomp,wang2025geak,wei2025astra,zhang2025cudaforge,liao2025kernelevolve,li2025tritonforge}. Their capacity upper bound is restricted by the foundational large language models. Robust benchmarking~\citep{lange2025towards} revealed that many such systems exploit testing loopholes such as eliminating redundant operators, hard-coding for specific input patterns. Our work focuses on training LLMs directly without introducing multi-turn agents, and is therefore complementary to these works.

\textbf{Synthetic Code Generation.\quad}
In the broader code generation landscape, methods like Magicoder~\citep{wei2023magicoder} and WizardCoder~\citep{luo2023wizardcoder} prompt LLMs to generate synthetic code for a given code snippet. This approach cannot guarantee the correctness and the coverage of the generated code due to the sampling nature of LLMs. Unlike prior work that relies on expensive LLM-based synthesis for generating kernel data~\citep{kernelbook2025,liao2025kernelevolve}, our framework uses the CSP-DAG algorithm to generate valid PyTorch programs with guaranteed correctness and uniform coverage, which aligns with classical program synthesis~\citep{gulwani2017program,ellis2021dreamcoder}.

\textbf{Reinforcement Learning with Verifiable Rewards (RLVR).\quad}
RLVR does not rely on human supervision, and has been shown to enable the emergence of new abilities or the discovery of knowledge beyond the original training data~\citep{guo2025deepseek}. 
In our work, we observe similar phenomena: although our training pipeline does not explicitly instruct the model on how to write Triton kernels, the model learns to map PyTorch code to Triton kernel code on its own—a new skill that naturally emerges through RLVR.

One problem in RLVR for code generation is sparse reward in the early stage of the training. AutoTriton~\citep{li2025autotriton} and TritonRL~\citep{woo2025tritonrl} use compiler feedback plus correctness of the generated kernels as a single verifiable reward. This does not address the sparsity of the reward. To tackle this challenge, we introduce curriculum learning with decoupled rewards. The idea of decoupled rewards was first proposed in DRPO~\citep{li2025drpo}. We show that the decoupled rewards greatly help us stabilize the training in the early stage, where the reward is sparse, accelerating the convergence speed of our training.

\section{CSP-DAG for PyTorch Program Generation}
\label{sec:generation}
Synthesizing PyTorch programs presents two challenges: (i) syntactic validity and (ii) satisfaction of tensor shape constraints across operators. For example, it is invalid to multiply two matrices if their dimensions do not match. To address these challenges, we propose an algorithm for synthesizing PyTorch programs with an arbitrary number of operators named  CSP-DAG. The algorithm consists of two stages: (1) randomly generating $N$ operators and linking them as a directed acyclic graph (DAG), and (2) finding the shapes of each operator's input tensors and output tensors via a CSP solver.

\subsection{Generating DAGs}

We represent a PyTorch program as a DAG where \textbf{edges} represent tensors with initially unknown shapes, and \textbf{nodes} represent operators that consume input edges and produce output edges. We assume each node may have multiple inputs but produces only one output.\footnote{An operator with two outputs, e.g., $a,b=f(x)$, can always be decomposed into two single-output operators: $a=f_1(x), b=f_2(x)$.}

Operators fall into two categories: \textbf{OpCompute} operators take tensor inputs and produce tensor outputs (e.g., \texttt{torch.add}, \texttt{torch.matmul}, \texttt{torch.relu}), while \textbf{OpCreate} operators produce tensors without any tensor inputs (e.g., \texttt{torch.randn}, \texttt{torch.ones}).
We list the full set of operators in Table~\ref{tab:operators} in Appendix~\ref{app:operators}. 

To generate a DAG, we maintain a list of candidate tensors, termed the candidate-tensor list, which is initialized to be an empty list. We iteratively add $n$ nodes (operators) into the graph. At each iteration,  we first randomly select an operator type $\texttt{op} \in \text{OpCompute}$. Let $m$ be the number of required inputs of $\texttt{op}$. Then, we randomly draw $m$ tensors without replacement from the candidate-tensor list to create edges attached to the node $\texttt{op}$. If the list has fewer than $m$ tensors, we randomly select $\texttt{op}' \in \text{OpCreate}$ and create a tensor with a line of code $\texttt{out} = \texttt{op'}(\texttt{shape})$ and insert the resulting tensor into the candidate-tensor list, whose shape will be determined later.  We repeat this step until the list has $m$ tensors. After the node $\texttt{op}$ and its $m$ edges $\text{in}_1,\text{in}_2\dots,\text{in}_m$ are created, we will create a line of code $\texttt{out}=\texttt{op}(\text{in}_1,\text{in}_2\dots,\text{in}_m)$ and insert $\texttt{out}$ into the candidate-tensor list. 

\begin{figure*}[h]
\centering
\includegraphics[width=\textwidth]{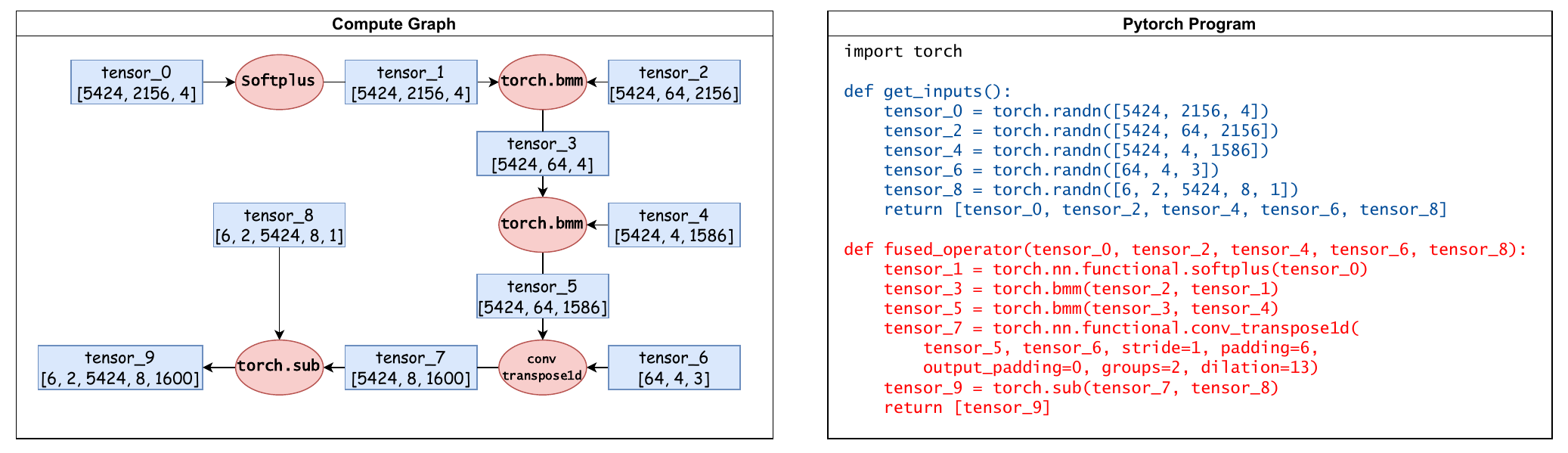}
\caption{Left: A DAG with 5 nodes and 10 edges. Each node represents an operator, and each edge represents a tensor. Right: The synthesized PyTorch program. Blue part indicates the tensors and the red part indicates the operators. }
\label{fig:dag}
\end{figure*}

We show an example of a DAG for a generated PyTorch program in Figure~\ref{fig:dag}(left).

\subsection{Constraint Programming} \label{sec:const_programming}
The above process generates valid computation graphs, but does not guarantee that tensor shapes are compatible across operators. For instance, matrix multiplication requires that the inner dimensions match. To generate valid PyTorch code, we must ensure all tensors satisfy the shape constraints imposed by the operators. The shape compatibility problem can be formulated as a constraint satisfaction problem (CSP) and solved using constraint programming tools like OR-Tools.

Formally, for each tensor edge $e$, we denote the order of this tensor by $n$ and the $i$-order has $s_i$ dimensions. Hence, each tensor edge has $n$ variables $s_1, \ldots, s_n$ to be solved. We restrict each dimension size to be an integer between $1$ and $2^{15}$. We let $e.s_i$ denote the variable $s_i$ attached to the tensor $e$. 

For each operator in the DAG, its order and dimension size are constrained by its neighbors in the DAG. Different operators impose different constraints, reflecting their underlying computational characteristics. We defer the full constraint specifications to Appendix~\ref{app:const}. 

In addition to operator-specific constraints, we further constrain the minimum and maximum FLOPs of the generated PyTorch module, the minimum and maximum number of elements of all tensors in the DAG:
\begin{align}
    \text{FLOPS}_\text{min} \leq \sum_{\text{op} \in \text{nodes}} \text{flops}(\text{op}) \leq \text{FLOPS}_\text{max}, \quad
    \text{SIZE}_\text{min} \leq \sum_{e \in \text{edges}} \text{numel}(e) \leq \text{SIZE}_\text{max},
\end{align}
where $\text{numel}(e) = \prod_{i=1}^{e.n} e.s_{i}$ is the total number of elements in the tensor $e$. The specific values of these bounds are provided in Appendix~\ref{app:const}.

The above inequalities form a typical constraint satisfaction problem (CSP). We use CP-SAT solver~\citep{cpsatlp} to find a feasible solution. If multiple feasible solutions exist, the CP-SAT solver selects one at random. Figure~\ref{fig:dag}(left) shows one solution of tensor shapes by the CP-SAT solver, whose corresponding PyTorch program is shown in Figure~\ref{fig:dag}(right). 

Our CSP-DAG algorithm guarantees coverage over all valid PyTorch programs composable from the given operators. Solving the constraints for a DAG of up to 20 operators takes \textbf{1--2 seconds} on a single CPU core. At scale, generating 100k programs on a 32-core machine requires roughly 1.5 hours in total.

\textbf{Difficulty Level.} In this work, we say that a PyTorch program is of difficulty level $k$ if it has $k$ OpCompute operators.

\textbf{Remark.} The format of our synthesized PyTorch programs closely resembles the intermediate representation (IR) used by PyTorch's compiler stack. In practice, most standard PyTorch models can be lowered into this format via \texttt{torch.export}~\citep{ansel2024pytorch}, which traces the computational graph and produces a flat, functional representation where each line corresponds to a single operator application.

\section{Training Pipeline} \label{sec:training}
In this section, we present our training pipeline based on the synthetic PyTorch programs. It consists of three training components — (i) SFT on a small set of PyTorch-Triton pairs on difficulty level 1; (ii) curriculum RL with decoupled rewards, and (iii) test-time search for optimal compositional kernels  — supported by a Triton verifier described in Section~\ref{sec:verifier}.

\subsection{Triton Verifier} \label{sec:verifier}
We consider a Triton translation of a PyTorch program to be ``correct'' if (i) it is syntactically valid, (ii) it passes the faithfulness validation, and (iii) it produces the same output as the PyTorch implementation.

{\bf Syntactic validation.} We apply a rule-based linter to confirm the presence of at least one \texttt{@triton.jit}-annotated kernel, then attempt compilation to catch remaining syntactic errors.

{\bf Faithfulness validation.} Sometimes the LLM may simply copy the PyTorch implementation, which of course generates the same outputs. To prevent this, we employ monkey-patch testing by replacing each kernel with a no-op kernel to verify that the Triton kernels are genuinely utilized.

{\bf Correctness validation.} We construct 5 random test cases with input-output pairs from the corresponding PyTorch program. We execute the Triton code on these inputs and compare its outputs with those produced by the PyTorch implementation using the same inputs. The test is passed only when all outputs of PyTorch against Triton are matched precisely.

\subsection{Supervised Fine-Tuning} \label{sec:sft}
We employ SFT to provide a cold start for RL. Since the base model may not possess sufficient knowledge of the Triton language, directly applying RL would lead to extremely sparse and uninformative rewards, as most generated programs would fail to compile.

We construct our SFT dataset using the synthetic PyTorch programs as described in Section~\ref{sec:generation}. Specifically, all SFT programs are \textbf{Level 1} programs (i.e., each containing exactly one OpCompute operator), which enables the model to first acquire basic PyTorch-to-Triton conversion skills before progressing to harder tasks in the RL stage. For each synthetic PyTorch program, we prompt DeepSeek-R1 or GPT-5.2 to generate the corresponding Triton kernel implementation. The Triton kernel is then verified, and we keep the Triton kernel codes that pass all verifications described in Section~\ref{sec:verifier}. Ultimately, this procedure yields a dataset of 2,026 PyTorch–Triton pairs covering 61 fundamental operators commonly used in PyTorch.
Detailed information about the dataset construction process is provided in Appendix~\ref{app:dataset}.

We denote these PyTorch-Triton pairs as $\mathcal D=\{\x_i, \y_i\}_{i=1}^{2026}$. We fine-tune a pretrained model by minimizing the SFT loss given below:
\begin{equation}
\min_{\theta} -\frac{1}{|\mathcal D|}\sum_{(\x,\y) \in \mathcal{D}} \log \pi_\theta(\y|\x),
\end{equation}
where $\pi_{\theta}$ denote the LLM with parameters $\theta$. 

\subsection{RLVR with Decoupled Rewards} \label{sec:rlvr}
To further enhance the capability of the model, we conduct RL after SFT. While most existing studies adopt GRPO as the workhorse, we employ DRPO~\citep{li2025drpo} which is designed to handle two decoupled reward signals: a correctness reward and a speed reward. 

The key idea of DRPO is to decouple the accuracy reward from the speed reward in the learning process. Let $q$ denote an input query, which is a PyTorch program to be translated, and let $o$ denote an output Triton implementation.  Let $s_\theta(o, q)$ denote the average log-likelihood score of the model: $s_\theta(o, q) = \frac{1}{|o|}\sum_{t=1}^{|o|} \log \pi_\theta(o_t|q, o_{<t})$, where $o_t$ is the $t$-th token of $o$. The goal of our RL is to (i) improve the log-likelihood of correct translations while decreasing that of incorrect ones; (ii) favor faster Triton implementations than slower ones. 

Let $\pi_{\text{old}}(\cdot)$ denote the old model to be improved. Let $\mathcal Q=\{q_1,\ldots, q_n\}$ denote a set of input queries. For each $q_i$, we let the model $\pi_{\text{old}}(\cdot)$ generate $m$ answers $\{o_{i,1}, \ldots, o_{i, m}\}$. We split these outputs into a correct set  $\S_+(q_i)$ and an incorrect set $\S_-(q_i)$ . For each output $o$, we denote its speed reward by $r_s(o|q)$. With these notations, the policy optimization of DRPO is formulated as a sum of three terms:
\begin{equation}
\begin{split}
\min_\theta  &- \frac{1}{n}\sum_{i=1}^n\sum_{o\in \mathcal S_+(q_i)} \omega(o|q)s_\theta(o,q) + \tau \log \Big(\frac{1}{|\mathcal S_-(q_i)|}\sum_{o'\in\mathcal S_-(q_i)}\exp\big(\frac{s_\theta(o',q)}{\tau}\big) \Big)  \\
&+ \beta\left[\mathbb{D}_{\text{KL}}(\pi_{\text{old}} || \pi_\theta) - \delta \right]_+^2,
\end{split}
\end{equation}
where $\omega(o|q)$ is a speed-induced weight for a correct output  defined by  $$\omega(o|q) = \frac{\exp(r_s(o|q)/\lambda)}{\sum_{o \in\mathcal S_+(q)} \exp(r_s(o|q)/\lambda)},$$ $\tau, \lambda, \delta, \beta>0$ are hyper-parameters. 

The first term is to increase the log-likelihood of correct outputs, with each correct output weighted using its speed reward. 
The second term is to decrease the log-likelihood of incorrect outputs. Using the log-sum-exp function also weights each incorrect output differently for the gradient calculation, i.e., a wrong output with a larger log-likelihood will get a higher weight for penalization. The last term is an adaptive regularization term, regularizing the change from the old model. It is motivated from the trust-region constraint $\mathbb{D}_{\text{KL}}(\pi_{\text{old}} || \pi_\theta)\leq \delta$ using a squared-hinge penalty function $\beta[\cdot]_+^2=\beta(0, \cdot)^2$. 

\textbf{Speed Reward.} We define the speed reward as a function of the ratio between the execution time of the PyTorch program and that of the corresponding Triton program $r_s(o|q) = f\left(\frac{t_{\text{torch}}}{t_{\text{triton}}}\right)$, where $t_{\text{torch}}$ and $t_{\text{triton}}$ are the execution time of the PyTorch and Triton implementations, respectively. In the experiments, we have evaluated different functions $f$ including the logarithmic function $f(s) = \log s$ and the power function $f(s)=s^\alpha$ with $\alpha>0$. The best choice is the logarithmic function. We present the ablation studies of speed reward functions in Section~\ref{sec:ab}.




\subsection{Curriculum Reinforcement Learning} \label{sec:curriculum-rl}
Curriculum learning has been widely adopted in reinforcement learning~\citep{10.5555/3455716.3455897,parashar2025curriculumreinforcementlearningeasy,wang2025dumpautomateddistributionlevelcurriculum}, as it improves learning efficiency and stability by gradually increasing task difficulty. We operationalize this by advancing to the next level once Pass@1 accuracy on held-out programs of the current level exceeds a threshold (50\% in our experiments). We stop advancing once performance plateaus, which occurs at Level~5 in our experiments.



\subsection{Test-time Search for Optimal Compositional Kernels}
For PyTorch programs too complex to fit in a single Triton kernel, we propose a test-time search that identifies the optimal composition of partial Triton kernels. To achieve this, we employ a systematic search procedure over different fusion strategies. Given a PyTorch program with $n$ sequential operators, our approach proceeds in three stages:

{\bf Fragment Extraction.} We decompose the program into all possible contiguous subsequences of operators with length $l\in \{1,2,\ldots, \min(5,n)\}$, where $n$ is the number of operators of the program. Fragments are extracted in order of increasing length --- first all single-operator fragments, then all two-operator fragments, and so on up to length 5. This generates $m = \sum_{l=1}^{\min(5,n)}(n-l+1) = O(n)$ candidate fragments, making the extraction cost linear in the program length and thus affordable even for complex programs.

{\bf Kernel Generation and Verification.} For each extracted fragment, we prompt our trained model to generate a corresponding Triton kernel implementation. We then verify the functional correctness of each generated kernel using the verification procedure described in Section~\ref{sec:verifier}. Let $m' \leq m$ denote the number of kernels that pass verification.

{\bf Code Reconstruction.} For each of the $m'$ verified kernels, we construct a candidate hybrid implementation by replacing the corresponding PyTorch fragment with the Triton kernel while keeping all other operators unchanged. This yields $m'$ distinct hybrid implementations. We benchmark each candidate by measuring its execution time and select the fastest implementation as our final output. This ensures we obtain the optimal fusion strategy among all verified options while maintaining correctness guarantees.

\section{Experiments} \label{sec:experiments}

\subsection{Experimental Setup}

We employ Qwen-2.5-Coder-7B-Instruct as our base model. Our training pipeline proceeds in two stages. First, SFT operates on 2,026 single-operator PyTorch-Triton pairs covering 61 fundamental operators (detailed in Section~\ref{sec:sft}), with ground-truth Triton implementations generated using DeepSeek-R1 or GPT-5.2. We train for 10 epochs with a learning rate $2 \times 10^{-6}$ and a batch size of 64. Second, we apply DRPO in three sequential curriculum stages: 20k programs at Level~1, 60k at Level~2, and 20k at Level~5. We use learning rate $1 \times 10^{-6}$, DRPO hyper-parameters $(\beta_0,\tau,\lambda) = (100,5,0.1)$, KL constraint upper bound $\delta = 0.001$, and generate 8 rollouts per prompt. All experiments are conducted on 8$\times$H100 GPUs. SFT takes approximately 2 hours; the full curriculum RL pipeline of 100k synthetic programs across three stages takes approximately 10 days.

\textbf{Benchmarks.} Our evaluation uses two complementary benchmarks. The synthetic benchmark contains 400 held-out PyTorch programs with no overlap with training data, spanning four difficulty levels: 100 single-operator programs (Level 1), 100 two-operator programs (Level 2), 100 five-operator programs (Level 5), and 100 twenty-operator programs (Level 20). This benchmark systematically evaluates the model's ability to scale from simple single-operator translation to complex multi-operator kernel fusion. The KernelBench benchmark comprises 250 PyTorch programs for neural network operations across three difficulty levels: Level 1 (100 tasks) contains single-kernel operators like convolution, matrix multiplications, and activations; Level 2 (100 tasks) includes fusion patterns such as convolution with bias and ReLU; and Level 3 (50 tasks) features complete ML architectures including MobileNet, VGG, and MiniGPT.

\textbf{Evaluations.} We employ the same verification protocol described in Section~\ref{sec:verifier} to verify the generated Triton kernels. We report three metrics: (1) Acc as the percentage of kernels that pass verification, (2) Faster1 as the percentage of kernels that are both correct and achieve more than 1x speedup over the PyTorch execution, and (3) average speedup over PyTorch, computed using the geometric mean. We report average speedup in two ways: over the verified kernels only (Success), and over all tasks with each failed conversion counted as a 1$\times$ speedup, i.e., falling back to the original PyTorch implementation (Full).

\textbf{Baselines.} We compare against several baselines spanning commercial LLMs (GPT-5.2, Claude Sonnet 4.5), open-source models (DeepSeek-R1, Qwen-3-Coder-480B), and a specialized model (AutoTriton). Since test-time search is model-agnostic, we additionally apply it to AutoTriton and, on KernelBench, to Claude-Sonnet-4.5. The prompt template is given in the Appendix~\ref{app:prompt}.

\begin{table}[t]
\centering
\setlength{\aboverulesep}{1.5pt}
\setlength{\belowrulesep}{1.5pt}
\renewcommand{\arraystretch}{0.85}
\setlength{\tabcolsep}{3pt}
\caption{Main results on synthetic benchmark. Acc (\%): Pass@1 accuracy breakdown by difficulty level. Faster1 (\%): percentage of correct kernels achieving $>1\times$ speedup. Avg. speedup: geometric mean speedup over PyTorch, computed over correct kernels only (Success) and over all tasks with failed conversions counted as $1\times$ (Full).}
\label{tab:main_results}
\footnotesize
\begin{tabular*}{\linewidth}{@{\extracolsep{\fill}}l*{10}{c}@{}}
\toprule
\multirow{2}{*}{Model} & \multicolumn{2}{c}{Level 1} & \multicolumn{2}{c}{Level 2} & \multicolumn{2}{c}{Level 5} & \multicolumn{2}{c}{Level 20} & \multicolumn{2}{c}{Avg. speedup} \\
\cmidrule(lr){2-3} \cmidrule(lr){4-5} \cmidrule(lr){6-7} \cmidrule(lr){8-9} \cmidrule(lr){10-11}
& Acc & Faster1 & Acc & Faster1 & Acc & Faster1 & Acc & Faster1 & Success & Full \\
\midrule
GPT-5.2 & 54 & 17 & 43 & 35 & 5 & 5 & 0 & 0 & 1.34 & 1.08 \\
Claude-Sonnet-4.5 & 68 & 14 & 49 & 29 & 7 & 3 & 0 & 0 & 0.68 & 0.89 \\
DeepSeek-R1 & 33 & 13 & 24 & 16 & 9 & 3 & 0 & 0 & 0.91 & 0.98 \\
Qwen-3-Coder-480B & 48 & 6 & 30 & 23 & 1 & 1 & 0 & 0 & 0.97 & 0.99 \\
AutoTriton & 36 & 7 & 15 & 8 & 2 & 1 & 0 & 0 & 1.08 & 1.01 \\
\quad + test-time search & 36 & 7 & 45 & 22 & 61 & 34 & 96 & 80 & 1.18 & 1.10 \\
\midrule
DRTriton (ours) & \textbf{87} & \textbf{32} & \textbf{75} & \textbf{54} & 15 & 10 & 0 & 0 & 1.20 & 1.08 \\
\quad + test-time search & \textbf{87} & \textbf{32} & \textbf{96} & \textbf{78} & \textbf{99} & \textbf{89} & \textbf{99} & \textbf{86} & \textbf{1.57} & \textbf{1.53} \\
\bottomrule
\end{tabular*}
\end{table}
\begin{figure}[t]
  \begin{center}
    \centerline{\includegraphics[width=\linewidth]{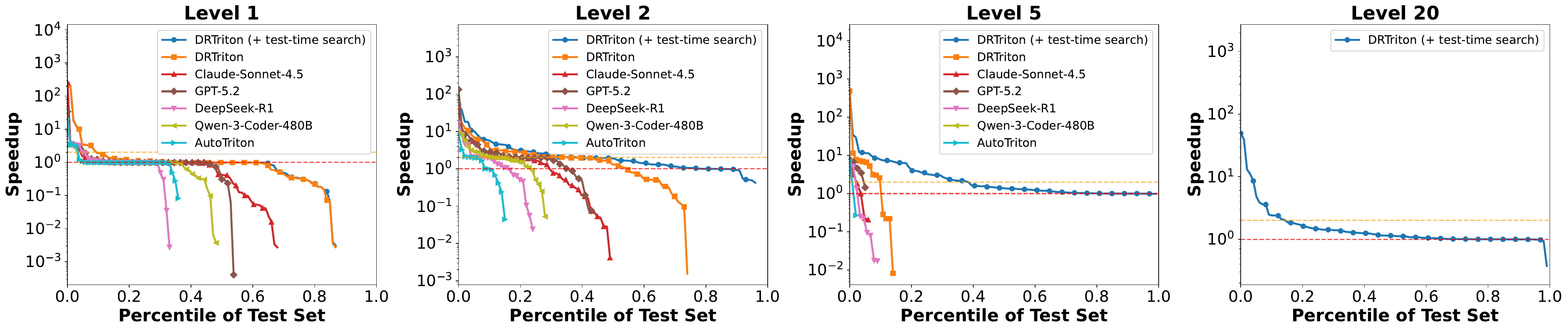}}
    \caption{Speedup distribution across difficulty levels for DRTriton, DRTriton (with test-time search), and other baselines, on synthetic benchmarks. Each curve shows speedups of generated Triton kernels over PyTorch programs across the percentages of testing data. The red dashed flat line is the reference line for 1x speedup, and the orange dashed flat line is the reference line for 2x speedup.}
    \label{fig:speedup}
  \end{center}
\end{figure}

\begin{table}[t]
\centering
\setlength{\aboverulesep}{1.5pt}
\setlength{\belowrulesep}{1.5pt}
\renewcommand{\arraystretch}{0.85}
\caption{Results on KernelBench real-world benchmark. We report accuracy (Acc) and Faster1 scores (percentage of correct kernels achieving $>1\times$ speedup) against two baselines: Torch Eager (TE) and \texttt{torch.compile} (TC).}
\label{tab:kernelbench}
\footnotesize
\begin{tabular*}{\linewidth}{@{\extracolsep{\fill}}l*{9}{c}@{}}
\toprule
\multirow{2}{*}{Model} & \multicolumn{3}{c}{Level 1} & \multicolumn{3}{c}{Level 2} & \multicolumn{3}{c}{Level 3} \\
\cmidrule(lr){2-4} \cmidrule(lr){5-7} \cmidrule(lr){8-10}
 & Acc & TE & TC & Acc & TE & TC & Acc & TE & TC \\
\midrule
AutoTriton & 44 & 12 & 6 & 45 & 14 & 12 & 32 & 14 & 6 \\
GPT-5.2 & 40 & 23 & 23 & 32 & 23 & 11 & 30 & 18 & 10 \\
Claude-Sonnet-4.5 & 46 & 8 & 8 & 36 & 19 & 12 & 36 & 12 & 12 \\
DeepSeek-R1 & 25 & 8 & 7 & 22 & 15 & 8 & 8 & 2 & 2 \\
Qwen-3-Coder-480B & 21 & 5 & 1 & 18 & 14 & 1 & 24 & 0 & 0 \\
\midrule
AutoTriton (test-time search) & 44 & 12 & 6 & 89 & 32 & 19 & \textbf{86} & 48 & 20 \\
Claude-Sonnet-4.5 (test-time search) & 67 & \textbf{41} & \textbf{30} & \textbf{96} & 68 & 42 & \textbf{86} & \textbf{64} & 30 \\
DRTriton (test-time search) & \textbf{69} & 17 & 12 & \textbf{96} & \textbf{92} & \textbf{56} & 76 & 54 & \textbf{34} \\
\bottomrule
\end{tabular*}
\end{table}

\begin{figure}[t]
  \centering
  \begin{subfigure}{0.66\linewidth}
    \includegraphics[width=\linewidth]{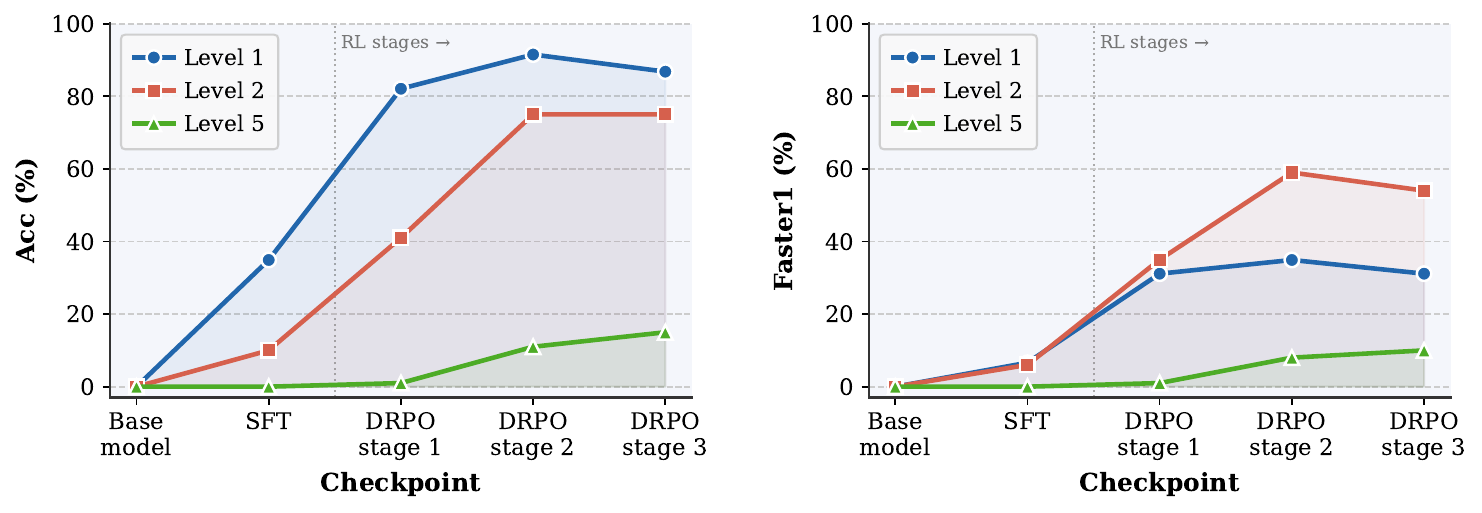}
    \caption{Curriculum RL progression.}
    \label{fig:progress}
  \end{subfigure}
  \hfill
  \begin{subfigure}{0.32\linewidth}
    \includegraphics[width=\linewidth]{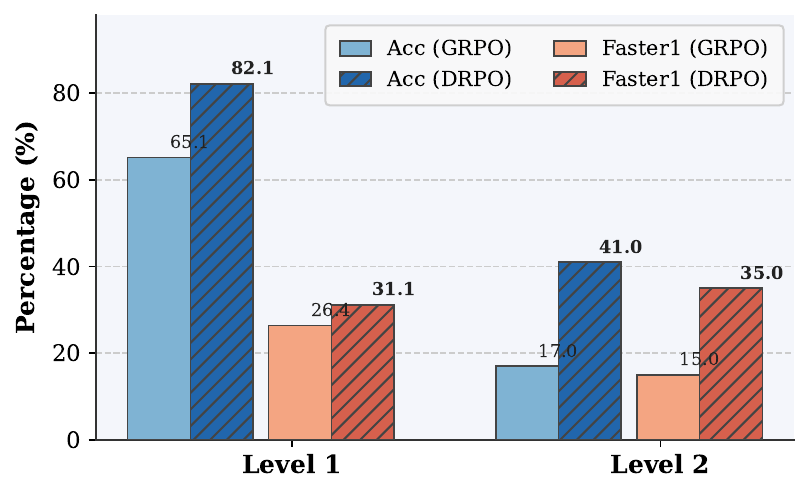}
    \vspace{0pt}
    \caption{DRPO vs.\ GRPO ablation.}
    \label{fig:drpo_vs_grpo}
  \end{subfigure}
  \caption{Training dynamics. (a) Performance progression across training stages on three difficulty levels, showing Acc and Faster1 metrics for the base model, after SFT, and across three DRPO stages. (b) Comparison of DRPO and GRPO trained on Stage 1 data (20k samples) from the same SFT checkpoint; DRPO consistently outperforms GRPO across all metrics.}
  \label{fig:training_dynamics}
\end{figure}

\subsection{Results on Synthetic Benchmarks}

Table~\ref{tab:main_results} and Figure~\ref{fig:speedup} present our main results on the synthetic benchmark, demonstrating that DRTriton substantially outperforms all baselines across all difficulty levels in both accuracy and speed.

{\bf DRTriton outperforms all baselines on accuracy.} DRTriton achieves 87\% accuracy at Level 1 (vs.\ 68\% for Claude) and 75\% at Level 2 (vs.\ 49\% for Claude). At Level 5, DRTriton achieves 15\% accuracy while most baselines fail almost completely.

{\bf DRTriton generates substantially faster kernels.} As shown in Figure~\ref{fig:speedup}, DRTriton's speedup curves lie clearly above all baselines across all difficulty levels. On Level 1, 32\% of DRTriton's kernels exceed 1$\times$ speedup, compared to 17\% for GPT-5.2 and 14\% for Claude; on Level 2 this rises to 54\%, far exceeding all baselines. This demonstrates that DRPO's speed-based reward weighting $\omega(o|q)$ successfully guides the model toward generating optimized, not merely correct, kernels.

{\bf Effectiveness of curriculum RL.} Figure~\ref{fig:progress} shows that DRPO training substantially improves over SFT. Peak performance at Level 1 and Level 2 is achieved at Stage 2, with a slight regression in Stage 3; Level 5 shows steady improvement throughout.

{\bf Effectiveness of test-time search.} Test-time search dramatically improves performance on complex programs. Counterintuitively, longer and harder programs achieve higher accuracy and better speedup after test-time search. This occurs because longer programs contain more code segments with optimization potential, making it easier for DRTriton to identify and improve inefficient fragments. Test-time search improves average speedup from 1.20$\times$ to 1.57$\times$ over correct kernels, and from 1.08$\times$ to 1.53$\times$ over the full benchmark, demonstrating that fragment-based kernel composition preserves and enhances optimization quality. Notably, test-time search is \textbf{model-agnostic}: as shown in Table~\ref{tab:main_results}, applying 
it to AutoTriton also yields substantial improvements across all difficulty levels. We further disentangle the contributions of the trained model and the search in Section~\ref{sec:ab}. The computational overhead of test-time search is reported in Appendix~\ref{app:tts_overhead}.

\subsection{Results on KernelBench}
While our synthesis framework generates programs in functional format, KernelBench provides object-oriented PyTorch code. To bridge this gap, we use \texttt{torch.export} to lower each KernelBench model into a flat functional IR, which has the same structure as our synthesized programs. Given the functional code, we then apply test-time search to identify the optimal kernel, and measure execution speed against both the original 
PyTorch implementation and the \texttt{torch.compile} baseline. An example of the rewritten code is provided in Appendix~\ref{app:rewriting}.

Table~\ref{tab:kernelbench} and Figure~\ref{fig:speedup_kernelbench} (Appendix~\ref{sec:spk}) present the results. Test-time search substantially improves every model it is applied to, and DRTriton with test-time search achieves substantially higher accuracy and faster execution than all baselines evaluated without search. At Level 1 (single operators), DRTriton achieves the highest accuracy (69\%) but modest speedups over Torch Eager (TE: 17\%) and \texttt{torch.compile} (TC: 12\%), as these operators are already well-optimized in PyTorch; here Claude-Sonnet-4.5 with test-time search reaches similar accuracy (67\%) with a larger fraction of faster kernels (TE: 41\%, TC: 30\%). At Level 2, DRTriton's performance improves dramatically: 96\% accuracy with 92\% faster than TE and 56\% faster than TC. Claude-Sonnet-4.5 with test-time search matches this accuracy, but DRTriton produces faster kernels on 24 and 14 more percentage points of tasks against TE and TC respectively, demonstrating DRTriton's strength in multi-operator fusion.

At Level 3 (complete architectures), DRTriton maintains 76\% accuracy with 54\% outperforming TE and 34\% outperforming TC, the highest TC score among all methods. AutoTriton and Claude-Sonnet-4.5 with test-time search achieve higher accuracy at Level 3 (86\%), and Claude-Sonnet-4.5 with search also produces more kernels faster than Torch Eager (64\%), while DRTriton remains ahead against \texttt{torch.compile} (34\% vs.\ 30\% and 20\%). We note that DRTriton attains this performance with a 7B open model. A detailed walkthrough of a Level 3 (LeNet-5) kernel generation — from object-oriented PyTorch to optimized Triton — is provided in Appendix~\ref{app:level3_case}.

\subsection{Ablation Studies}\label{sec:ab}


We conduct three ablation studies to validate key design choices and to separate the contributions of the trained model and test-time search.

\textbf{Ablation on RL algorithm.} Figure~\ref{fig:drpo_vs_grpo} compares DRPO with GRPO on identical Stage 1 training data from the same SFT checkpoint. For GRPO, we design the reward as $r(o)=  1 + f\left(\frac{t_{\text{torch}}}{t_{\text{triton}}}\right)$ if the output is valid, otherwise, $r(o)= 0$. The GRPO objective is detailed in Appendix~\ref{app:GRPO}. DRPO consistently outperforms GRPO across all metrics.

\textbf{Ablation on reward design.} Table~\ref{tab:reward_ablation} in Appendix~\ref{app:results} examines different reward function designs for the speed component $r_s(o)$ in DRPO. We compare the logarithmic form $r_s(o) = \log(t_{\text{torch}}/t_{\text{triton}})$ against power forms $r_s(o) = (t_{\text{torch}}/t_{\text{triton}})^\alpha$ for $\alpha \in \{0.25, 0.5, 0.75, 1.0\}$. The logarithmic form consistently achieves the best Acc and Faster1, and is used in all other experiments.

\begin{table}[t]
\centering
\setlength{\aboverulesep}{1.5pt}
\setlength{\belowrulesep}{1.5pt}
\renewcommand{\arraystretch}{0.85}
\setlength{\tabcolsep}{3pt}
\caption{Decomposing the contributions of the model and test-time search on the synthetic benchmark. The same test-time search is applied to the base model (Qwen-2.5-Coder-7B-Instruct), the SFT-only checkpoint, and the checkpoint after each DRPO stage. Metrics are defined as in Table~\ref{tab:main_results}.}
\label{tab:decomposition}
\footnotesize
\begin{tabular*}{\linewidth}{@{\extracolsep{\fill}}l*{10}{c}@{}}
\toprule
\multirow{2}{*}{Checkpoint (+ search)} & \multicolumn{2}{c}{Level 1} & \multicolumn{2}{c}{Level 2} & \multicolumn{2}{c}{Level 5} & \multicolumn{2}{c}{Level 20} & \multicolumn{2}{c}{Avg. speedup} \\
\cmidrule(lr){2-3} \cmidrule(lr){4-5} \cmidrule(lr){6-7} \cmidrule(lr){8-9} \cmidrule(lr){10-11}
& Acc & Faster1 & Acc & Faster1 & Acc & Faster1 & Acc & Faster1 & Success & Full \\
\midrule
Base model & 23 & 4 & 44 & 12 & 73 & 34 & 81 & 52 & 1.06 & 1.03 \\
SFT-only & 30 & 6 & 50 & 14 & 82 & 44 & 95 & 71 & 1.09 & 1.05 \\
DRPO Stage 1 & 82 & 28 & 92 & 62 & \textbf{100} & 89 & 98 & \textbf{87} & 1.32 & 1.29 \\
DRPO Stage 2 & \textbf{92} & \textbf{36} & \textbf{98} & \textbf{83} & \textbf{100} & \textbf{91} & 97 & 84 & 1.53 & 1.51 \\
DRPO Stage 3 (DRTriton) & 87 & 32 & 96 & 78 & 99 & 89 & \textbf{99} & 86 & \textbf{1.57} & \textbf{1.53} \\
\bottomrule
\end{tabular*}
\end{table}

\textbf{Contributions of the model versus search.} To disentangle the gains due to the trained model from those due to test-time search, we apply the same test-time search to the base model, the SFT-only checkpoint, and the checkpoint after each DRPO stage (Table~\ref{tab:decomposition}). Search alone is effective at raising accuracy on long programs: even the base model reaches 81\% accuracy at Level 20, because it only needs to succeed on some fragments. However, search contributes little to speed on its own: the base model and SFT-only checkpoint reach full-benchmark average speedups of only 1.03$\times$ and 1.05$\times$, with Level 1 and Level 2 Faster1 at or below 14\%. The largest gain comes from the first DRPO stage, which raises the full-benchmark average speedup from 1.05$\times$ to 1.29$\times$ and Level 2 Faster1 from 14\% to 62\%. Subsequent stages further increase speed (1.51$\times$ and 1.53$\times$). Consistent with Figure~\ref{fig:progress}, Stage 2 gives the best Level 1 and Level 2 results, while Stage 3 gives the highest average speedup. These results indicate that test-time search mainly improves coverage, while the speed of the final kernels is driven primarily by the RL-trained model.

\section{Conclusion}
We have presented DRTriton, a scalable framework for training PyTorch-to-Triton LLMs via large-scale synthetic data and curriculum reinforcement learning. DRTriton-7B outperforms state-of-the-art commercial LLMs and specialized models on the synthetic benchmark, and on KernelBench it achieves the highest accuracy at Levels 1--2 and the most kernels faster than \texttt{torch.compile} at Levels 2--3, remaining competitive with a frontier commercial model equipped with the same test-time search despite having only 7B parameters. These results demonstrate that a small SFT warmup on synthetic data combined with RL on large-scale synthetic programs generalizes effectively to complex, human-authored GPU kernels. Our current operator set is limited to 61 PyTorch operators, and extending coverage to sparse operations, custom CUDA extensions, and native CUDA generation is an important future direction. We believe this work establishes synthetic data-driven curriculum RL as a scalable and promising paradigm for automated GPU kernel generation.

\begin{ack}
S. Guo and T. Yang were partially supported by National Science Foundation Award 2306572. Part of the GPU resources used in this work were provided by Oracle. We thank Ming Lin for facilitating access to these resources and for assistance in running experiments.
\end{ack}

\bibliographystyle{plainnat}
\bibliography{main}


\newpage
\appendix

\section{Full set of operators} \label{app:operators}
{
\small
\begin{longtable}{>{\raggedright\arraybackslash}p{3cm}>{\raggedright\arraybackslash}p{2.5cm}>{\raggedright\arraybackslash}p{6.5cm}>{\centering\arraybackslash}p{1cm}}
\caption{Complete List of Operators} \label{tab:operators} \\
\toprule
\textbf{Category} & \textbf{Operator Class} & \textbf{PyTorch Function} & \textbf{\#In} \\
\midrule
\endfirsthead

\multicolumn{4}{c}{\tablename\ \thetable\ -- \textit{Continued from previous page}} \\
\toprule
\textbf{Category} & \textbf{Operator Class} & \textbf{PyTorch Function} & \textbf{\#In} \\
\midrule
\endhead

\midrule
\multicolumn{4}{r}{\textit{Continued on next page}} \\
\endfoot

\bottomrule
\endlastfoot

\multirow{3}{3cm}{\textbf{OpCreate}} & \texttt{Randn} & \texttt{torch.randn} & 0 \\
& \texttt{Ones} & \texttt{torch.ones} & 0 \\
& \texttt{Zeros} & \texttt{torch.zeros} & 0 \\
\midrule

\multirow{6}{3cm}{\textbf{Elementwise Binary}} & \texttt{Add} & \texttt{torch.add} & 2 \\
& \texttt{Mul} & \texttt{torch.mul} & 2 \\
& \texttt{Sub} & \texttt{torch.sub} & 2 \\
& \texttt{Div} & \texttt{torch.div} & 2 \\
& \texttt{Maximum} & \texttt{torch.maximum} & 2 \\
& \texttt{Minimum} & \texttt{torch.minimum} & 2 \\
\midrule

\multirow{1}{3cm}{\textbf{Elementwise Ternary}} & \texttt{Lerp} & \texttt{torch.lerp} & 3 \\
\midrule

\multirow{8}{3cm}{\textbf{Reduction}} & \texttt{Max} & \texttt{torch.max} & 1 \\
& \texttt{Min} & \texttt{torch.min} & 1 \\
& \texttt{Sum} & \texttt{torch.sum} & 1 \\
& \texttt{Mean} & \texttt{torch.mean} & 1 \\
& \texttt{ArgMax} & \texttt{torch.argmax} & 1 \\
& \texttt{ArgMin} & \texttt{torch.argmin} & 1 \\
& \texttt{Var} & \texttt{torch.var} & 1 \\
& \texttt{Norm} & \texttt{torch.norm} & 1 \\
\midrule

\multirow{5}{3cm}{\textbf{Matrix Operators}} & \texttt{Matmul} & \texttt{torch.matmul} & 2 \\
& \texttt{Bmm} & \texttt{torch.bmm} & 2 \\
& \texttt{Transpose} & \texttt{torch.transpose} & 1 \\
& \texttt{Triu} & \texttt{torch.triu} & 1 \\
& \texttt{Tril} & \texttt{torch.tril} & 1 \\
\midrule
\multirow{14}{3cm}{\textbf{Unary Activation}} & \texttt{ReLU} & \texttt{torch.relu} & 1 \\
& \texttt{LeakyReLU} & \texttt{torch.nn.functional.leaky\_relu} & 1 \\
& \texttt{Sigmoid} & \texttt{torch.sigmoid} & 1 \\
& \texttt{Tanh} & \texttt{torch.tanh} & 1 \\
& \texttt{Swish} & \texttt{torch.nn.functional.silu} & 1 \\
& \texttt{GELU} & \texttt{torch.nn.functional.gelu} & 1 \\
& \texttt{SELU} & \texttt{torch.selu} & 1 \\
& \texttt{ELU} & \texttt{torch.nn.functional.elu} & 1 \\
& \texttt{Hardsigmoid} & \texttt{torch.nn.functional.hardsigmoid} & 1 \\
& \texttt{HardTanh} & \texttt{torch.nn.functional.hardtanh} & 1 \\
& \texttt{Softplus} & \texttt{torch.nn.functional.softplus} & 1 \\
& \texttt{Softsign} & \texttt{torch.nn.functional.softsign} & 1 \\
& \texttt{LogSigmoid} & \texttt{torch.nn.functional.logsigmoid} & 1 \\
& \texttt{Clamp} & \texttt{torch.clamp} & 1 \\
\midrule
\multirow{2}{3cm}{\textbf{Unary with Dimension}} & \texttt{Softmax} & \texttt{torch.softmax} & 1 \\
& \texttt{LogSoftmax} & \texttt{torch.log\_softmax} & 1 \\
\midrule

\multirow{4}{3cm}{\textbf{Unary Mathematical}} & \texttt{Cos} & \texttt{torch.cos} & 1 \\
& \texttt{Sin} & \texttt{torch.sin} & 1 \\
& \texttt{Exp2} & \texttt{torch.exp2} & 1 \\
& \texttt{Abs} & \texttt{torch.abs} & 1 \\
\midrule

\multirow{3}{3cm}{\textbf{Cumulative Operators}} & \texttt{CumMax} & \texttt{torch.cummax} & 1 \\
& \texttt{CumMin} & \texttt{torch.cummin} & 1 \\
& \texttt{CumSum} & \texttt{torch.cumsum} & 1 \\
\midrule

\multirow{4}{3cm}{\textbf{Normalization}} & \texttt{BatchNorm} & \texttt{torch.nn.functional.batch\_norm} & 1 \\
& \texttt{LayerNorm} & \texttt{torch.nn.functional.layer\_norm} & 1 \\
& \texttt{GroupNorm} & \texttt{torch.nn.functional.group\_norm} & 1 \\
& \texttt{InstanceNorm} & \texttt{torch.nn.functional.instance\_norm} & 1 \\
\midrule

\multirow{2}{3cm}{\textbf{Pooling (1D)}} & \texttt{AvgPool1d} & \texttt{torch.nn.functional.avg\_pool1d} & 1 \\
& \texttt{MaxPool1d} & \texttt{torch.nn.functional.max\_pool1d} & 1 \\
\midrule

\multirow{2}{3cm}{\textbf{Pooling (2D)}} & \texttt{AvgPool2d} & \texttt{torch.nn.functional.avg\_pool2d} & 1 \\
& \texttt{MaxPool2d} & \texttt{torch.nn.functional.max\_pool2d} & 1 \\
\midrule

\multirow{2}{3cm}{\textbf{Pooling (3D)}} & \texttt{AvgPool3d} & \texttt{torch.nn.functional.avg\_pool3d} & 1 \\
& \texttt{MaxPool3d} & \texttt{torch.nn.functional.max\_pool3d} & 1 \\
\midrule

\multirow{3}{3cm}{\textbf{Convolution}} & \texttt{Conv1d} & \texttt{torch.nn.functional.conv1d} & 2 \\
& \texttt{Conv2d} & \texttt{torch.nn.functional.conv2d} & 2 \\
& \texttt{Conv3d} & \texttt{torch.nn.functional.conv3d} & 2 \\
\midrule

\multirow{3}{3cm}{\textbf{Transposed Convolution}} & \texttt{ConvTranspose1d} & \texttt{torch.nn.functional.conv\_transpose1d} & 2 \\
& \texttt{ConvTranspose2d} & \texttt{torch.nn.functional.conv\_transpose2d} & 2 \\
& \texttt{ConvTranspose3d} & \texttt{torch.nn.functional.conv\_transpose3d} & 2 \\
\midrule

\multirow{2}{3cm}{\textbf{Tensor Manipulation}} & \texttt{Cat} & \texttt{torch.cat} & $\geq$2 \\
& \texttt{Stack} & \texttt{torch.stack} & $\geq$2 \\

\end{longtable}
}

\section{Constraint Details} \label{app:const}

Before describing the constraints for each operator category, we first clarify the broadcasting mechanism and our notation convention. \textbf{Broadcasting} is a mechanism that allows operators to work with inputs of different shapes by automatically expanding dimensions. Specifically, two dimensions are compatible for broadcasting if they are equal, or if one of them is 1. When broadcasting occurs, the dimension of size 1 is virtually replicated to match the larger dimension. For example, consider adding two tensors: tensor $a$ with shape $(3, 1, 5)$ and tensor $b$ with shape $(3, 4, 5)$. These are compatible for broadcasting because in each dimension position (comparing from right to left), we have: dimension 3 has $5 = 5$ (compatible), dimension 2 has $1$ and $4$ (compatible, 1 broadcasts to 4), and dimension 1 has $3 = 3$ (compatible). The output has shape $(3, 4, 5)$. However, tensor $a$ with shape $(3, 2, 5)$ and tensor $b$ with shape $(3, 4, 5)$ would fail to broadcast because dimension 2 has sizes $2$ and $4$, where neither is equal nor is either 1. Broadcasting also works across tensors with different orders (number of dimensions). For instance, tensor $a$ with shape $(5,)$ and tensor $b$ with shape $(3, 4, 5)$ can be broadcast together. Aligning from the right, $a$ is treated as having shape $(1, 1, 5)$, which is compatible with $(3, 4, 5)$, yielding an output of shape $(3, 4, 5)$.

Since constraint checking always occurs on the trailing (rightmost) dimensions of input tensors, and operators can have multiple inputs with different orders, we introduce a global variable $N = \max(\text{in}_1.n, \text{in}_2.n, \dots, \text{in}_m.n)$ representing the maximum order among all $m$ inputs. We then define right-aligned versions $\text{in}_1', \text{in}_2', \dots, \text{in}_m'$, each of order $N$, where $\text{in}_x'.s_i = \text{in}_x.s_{i-(N-\text{in}_x.n)}$ for all $i \in \{N-\text{in}_x.n+1, \dots, N\}$, and $\text{in}_x'.s_i = 1$ for all $i \in \{1, \dots, N-\text{in}_x.n\}$. This right-alignment pads leading dimensions with size 1, enabling uniform constraint specification across inputs of varying orders. Unless explicitly stated otherwise, all constraints below are defined on these right-aligned variables $\text{in}_1', \text{in}_2', \dots, \text{in}_m'$ rather than on the original inputs.

\begin{enumerate}
    \item \textbf{ElementwiseOp.} This category of operators takes two inputs and produces one output $c = \text{op}(a,b)$. The shapes of inputs can be broadcast. The constraints are:
    \begin{align*}
        & c.n = \max(a.n, b.n) \\
        & c.s_i = \max(a.s_i,b.s_i), \quad \forall i \in \{1, \dots, N\}\\
        & (a.s_i = b.s_i) \vee (a.s_i = 1) \vee (b.s_i = 1), \\ & \quad \forall i \in \{1, \dots, N\}
    \end{align*}
    
    \item \textbf{ReduceOp.} This category of operators takes one input and arguments dim and keepdim. Let $a$ denote the input, $c$ denote the output, $\text{dim}'$ denote the actual dimension index after right-alignment: $\text{dim}' = N - a.n + \text{dim} + 1$. The constraints are:
    \begin{align*}
        & \text{dim} < a.n \\
        & \text{if } \text{keepdim} = \text{True}: \notag \\
        & \quad c.n = a.n \\
        & \quad c.s_i = \begin{cases}
            a.s_i & \text{if } i \neq \text{dim}' \\
            1 & \text{if } i = \text{dim}'
        \end{cases} \\
        & \text{if } \text{keepdim} = \text{False}: \notag \\
        & \quad c.n = a.n - 1 \\
        & \quad c.s_i = \begin{cases}
            a.s_{i+1} & \text{if } i < \text{dim}' \\
            a.s_i & \text{if } i > \text{dim}'
        \end{cases} \\
        & \forall i \in \{1, \dots, N\}
    \end{align*}
    
    \item \textbf{Matmul.} Matrix multiplication with broadcasting on batch dimensions. Let $a$ and $b$ denote the inputs and $c$ denote the output:
    \begin{align*}
        & a.n \geq 2, \quad b.n \geq 2 \\
        & a.s_{N} = b.s_{N-1} \\
        & c.s_{N-1} = a.s_{N-1} \\
        & c.s_{N} = b.s_{N} \\
        & c.n = \max(a.n, b.n) \\
        & c.s_i = \max(a.s_i,b.s_i), \quad \forall i \in \{1, \dots, N-2\} \\
        & (a.s_i = b.s_i) \vee (a.s_i = 1) \vee (b.s_i = 1), \\
        & \quad \forall i \in \{1, \dots, N-2\} \\
    \end{align*}
    
    \item \textbf{Transpose.} Let $a$ denote the input, $c$ denote the output. Transposing the last two dimensions has the constraints:
    \begin{align*}
        & a.n \geq 2 \\
        & c.n = a.n \\
        & c.s_i = a.s_i, \quad \forall i \in \{1, \dots, N-2\} \\
        & c.s_{N-1} = a.s_{N} \\
        & c.s_{N} = a.s_{N-1}
    \end{align*}
    
    \item \textbf{ConvNd.} Convolution with parameters spatial dims $m \in \{1,2,3\}$, output channels $C_{\text{out}}$, kernel sizes $k_1, \dots, k_m$ for each spatial dimension, stride $s$, padding $p$, dilation $d$, and groups $g$:
    \begin{align*}
        & a.n = c.n = m + 2 \\
        & 2p \leq \max_j k_j \\
        & c.s_{N - m - 1} = a.s_{N - m - 1} \\
        & a.s_{N - m} \equiv 0 \pmod{g} \\
        & c.s_{N - m} = C_{\text{out}} \equiv 0 \pmod{g} \\
        & \text{For spatial dimension } i \in \{N - m + 1, \dots, N\}: \notag \\
        & \quad a.s_i \geq k_{i-(N-m)} \\
        & \quad c.s_i = \left\lfloor \frac{a.s_i + 2p - d(k_{i-(N-m)} - 1) - 1}{s} \right\rfloor + 1
    \end{align*}

    Other operators like PoolNd, ConvTransposeNd are handled similarly to ConvNd.
    
    \item \textbf{BatchNorm/LayerNorm/GroupNorm/InstanceNorm.} Normalization operators preserve the input shape:
    \begin{align*}
        & c.n = a.n \\
        & c.s_i = a.s_i, \quad \forall i \in \{1, \dots, N\}
    \end{align*}
    For specific normalization types, additional constraints may apply (e.g., GroupNorm requires the number of channels to be divisible by the number of groups).
\end{enumerate}

\textbf{FLOPs and Size Constraints.} In our experiments, we set $\text{FLOPS}_\text{min} = 2^{34}$, $\text{FLOPS}_\text{max} = 2^{35}$, $\text{SIZE}_\text{min} = 32$, and $\text{SIZE}_\text{max} = 2^{32}$.

\section{SFT Dataset Construction Details} \label{app:dataset}

Our dataset construction follows a multi-stage process designed to ensure both coverage and quality across diverse PyTorch operators.


{\bf Initial Data Generation.} We begin by sampling 200 synthetic PyTorch programs for each of the 61 fundamental operators, yielding 12,200 programs in total. For each program, we prompt DeepSeek-R1 to generate the corresponding Triton kernel implementation. We then validate each generated kernel through execution testing to verify functional correctness. Programs for which DeepSeek-R1 fails to produce correct Triton kernels are discarded. This initial filtering process results in 1,464 valid PyTorch–Triton pairs.

{\bf Iterative Refinement.} To identify operators requiring additional training data, we train an initial model using the 1,464-sample dataset through supervised fine-tuning (SFT) followed by one stage of reinforcement learning (RL). We then evaluate the trained model's operator-level success rate by generating and validating Triton kernels for 100 test samples per operator. This analysis reveals that certain operators exhibit extremely low success rates, indicating insufficient training data coverage. For operators with poor performance in the evaluation phase, we generate additional training data using GPT-5.2. Specifically, we prompt GPT-5.2 to produce 100 additional PyTorch–Triton pairs for each under-performing operator. After validation and filtering, this augmentation yields 562 additional correct pairs.

{\bf Final Dataset.} We merge the initial 1,464 samples with the 562 augmented samples to form our final training dataset of 2,026 PyTorch–Triton pairs. This dataset maintains broad coverage across all 61 operators while providing enhanced representation for operators that proved challenging for the base model. The distribution of operators in the final dataset reflects both the initial uniform sampling strategy and the targeted augmentation based on empirical performance needs.

\section{Additional Experiment Results}

\subsection{Speedup Curves on KernelBench}\label{sec:spk}

\begin{figure*}[ht]
  \begin{center}
    \centerline{\includegraphics[width=0.95\linewidth]{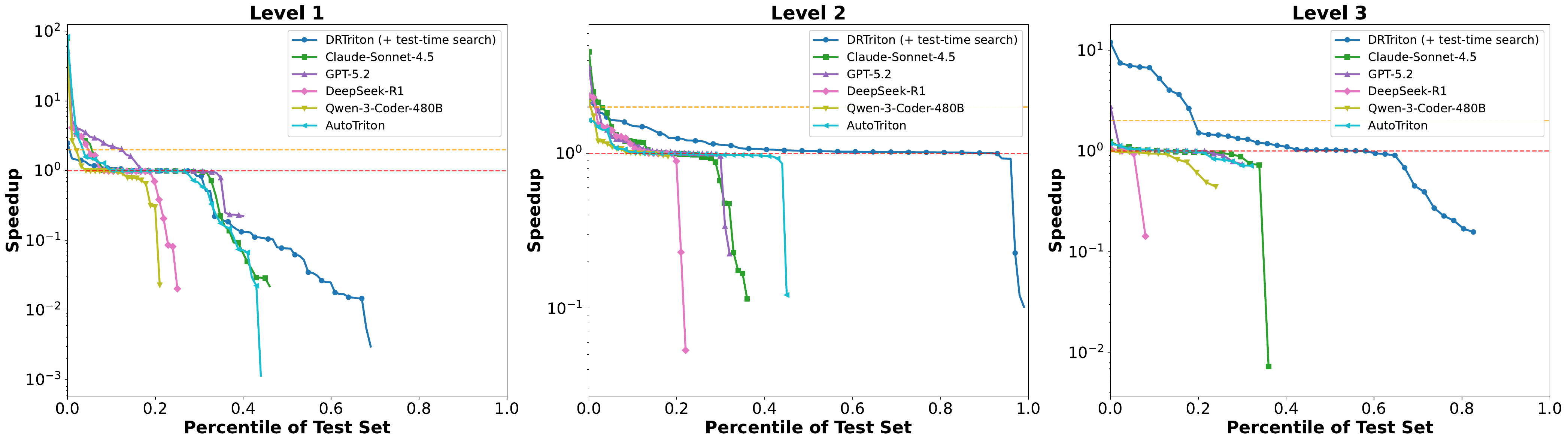}}
    \caption{Speedup distribution across difficulty levels for DRTriton (with test-time search), and other baselines, on KernelBench. Each curve shows speedups of generated Triton kernels over PyTorch programs across the percentages of testing data. The red dashed flat line is the reference line for 1x speedup, and the orange dashed flat line is the reference line for 2x speedup.}
    \label{fig:speedup_kernelbench}
  \end{center}
\end{figure*}

Figure~\ref{fig:speedup_kernelbench} shows the speedup distribution across 
all three KernelBench difficulty levels for DRTriton (with test-time search) 
and all baseline models.

\subsection{Ablation on Speed Reward Function} \label{app:results}

Table~\ref{tab:reward_ablation} compares different functional forms for the speed reward component $r_s(o)$ in DRPO, using Qwen-2.5-Coder-1.5B as the base model.

\begin{table}[th]
\centering
\caption{Ablation study on speed reward function design. All models are trained with DRPO on Stage 1 data using Qwen-2.5-Coder-1.5B as the base model, starting from the SFT checkpoint.}
\label{tab:reward_ablation}
\small
\begin{tabular}{@{}l*{4}{c}@{}}
\toprule
\multirow{2}{*}{Reward Function} & \multicolumn{2}{c}{Level 1}  \\
\cmidrule(lr){2-3} 
& Acc & Faster1  \\
\midrule
Power ($\alpha=0.25$) & 38.1 & 14.4  \\
Power ($\alpha=0.50$) & 30.9 & 5.2   \\
Power ($\alpha=0.75$) & 30.9 & 2.1  \\
Power ($\alpha=1.0$) & 32.0 & 3.1   \\
Logarithmic & \textbf{42.3} & \textbf{18.6} \\
\bottomrule
\end{tabular}
\end{table}

\subsection{Test-Time Search Overhead}
\label{app:tts_overhead}

Table~\ref{tab:tts_overhead} reports the wall-clock cost of test-time search on both benchmarks. Despite the larger number of fragments evaluated, the total overhead remains modest in absolute terms and is negligible relative to the overall training cost.

\begin{table}[h]
\centering
\caption{Test-time search overhead on synthetic and KernelBench benchmarks. Generation and validation times are reported in hours.}
\label{tab:tts_overhead}
\small
\begin{tabular}{@{}llccc c@{}}
\toprule
Benchmark & Setting & Fragments evaluated & Generation (h) & Validation (h) & Total (h) \\
\midrule
\multirow{2}{*}{Synthetic} & w/o test-time search & 400 & 0.01 & 0.11 & 0.12 \\
 & w/ test-time search & 10,612 & 0.13 & 2.72 & 2.85 \\
\midrule
\multirow{2}{*}{KernelBench} & w/o test-time search & 250 & 0.01 & 0.07 & 0.08 \\
 & w/ test-time search & 16,859 & 0.23 & 3.30 & 3.53 \\
\bottomrule
\end{tabular}
\end{table}

\section{PyTorch Code Rewriting Example} \label{app:rewriting}

The following shows a typical multi-layer perceptron (MLP) implementation from KernelBench:

\begin{lstlisting}[language=Python, basicstyle=\ttfamily\scriptsize]
import torch
import torch.nn as nn
import torch.nn.functional as F

class Model(nn.Module):
    def __init__(self, input_size, layer_sizes, output_size):
        """
        :param input_size: The number of input features
        :param layer_sizes: A list of ints containing the sizes 
                           of each hidden layer
        :param output_size: The number of output features
        """
        super(Model, self).__init__()
        
        layers = []
        current_input_size = input_size
        
        for layer_size in layer_sizes:
            layers.append(nn.Linear(current_input_size, layer_size))
            layers.append(nn.ReLU())
            current_input_size = layer_size
        
        layers.append(nn.Linear(current_input_size, output_size))
        
        self.network = nn.Sequential(*layers)
    
    def forward(self, x):
        """
        :param x: The input tensor, shape (batch_size, input_size)
        :return: The output tensor, shape (batch_size, output_size)
        """
        return self.network(x)

# Test configuration
batch_size = 128
input_size = 16384
layer_sizes = [16384, 16384]
output_size = 8192

def get_inputs():
    return [torch.rand(batch_size, input_size)]

def get_init_inputs():
    return [input_size, layer_sizes, output_size]
\end{lstlisting}

Using \texttt{torch.export} to trace the computational graph, we obtain the following functional representation. The rewritten code explicitly materializes all model parameters (weights and biases) and expresses the forward pass as a sequence of functional operators in \texttt{fused\_operator}. This functional representation aligns with our training data distribution and enables DRTriton to identify optimization opportunities.

\begin{lstlisting}[language=Python, basicstyle=\ttfamily\scriptsize]
import torch
import torch.nn as nn
import torch.nn.functional as F
import math

def get_inputs():
    # Initialize weights and biases for first linear layer
    tensor_0 = torch.empty((16384, 16384), device=None, dtype=None)
    nn.init.kaiming_uniform_(tensor_0, a=2.23606797749979)
    tensor_1 = torch.empty(16384, device=None, dtype=None)
    nn.init.uniform_(tensor_1, -0.0078125, 0.0078125)
    
    # Initialize weights and biases for second linear layer
    tensor_2 = torch.empty((16384, 16384), device=None, dtype=None)
    nn.init.kaiming_uniform_(tensor_2, a=2.23606797749979)
    tensor_3 = torch.empty(16384, device=None, dtype=None)
    nn.init.uniform_(tensor_3, -0.0078125, 0.0078125)
    
    # Initialize weights and biases for output layer
    tensor_4 = torch.empty((8192, 16384), device=None, dtype=None)
    nn.init.kaiming_uniform_(tensor_4, a=2.23606797749979)
    tensor_5 = torch.empty(8192, device=None, dtype=None)
    nn.init.uniform_(tensor_5, -0.0078125, 0.0078125)
    
    # Input tensor
    tensor_6 = torch.rand(128, 16384)

    return [tensor_6, tensor_0, tensor_1, tensor_2, tensor_3, 
            tensor_4, tensor_5]

def fused_operator(tensor_6, tensor_0, tensor_1, tensor_2, 
                     tensor_3, tensor_4, tensor_5):
    # First linear layer + ReLU
    tensor_20 = F.linear(tensor_6, tensor_0, tensor_1)
    tensor_21 = F.relu(tensor_20, inplace=False)
    
    # Second linear layer + ReLU
    tensor_22 = F.linear(tensor_21, tensor_2, tensor_3)
    tensor_23 = F.relu(tensor_22, inplace=False)
    
    # Output layer
    tensor_24 = F.linear(tensor_23, tensor_4, tensor_5)

    return [tensor_24]
\end{lstlisting}

\section{Evaluation Prompts} \label{app:prompt}
For KernelBench evaluation, we use the Triton backend and employ the default prompt provided by the KernelBench pipeline.

For evaluation on our synthetic benchmarks, we use the prompt template shown in the listing below. The prompt specifies detailed requirements for kernel structure, memory efficiency, and performance optimization, and includes a concrete example to guide code generation.

\begin{lstlisting}[basicstyle=\ttfamily\scriptsize]
## Task: Convert PyTorch to Efficient Triton Kernel

Convert the PyTorch `fused_operator()` function into an **efficient** 
Triton GPU kernel that produces identical numerical results while 
maximizing performance.

## Requirements

1. **Kernel Structure**: Use `@triton.jit` on **exactly one** kernel function; create entrypoint `triton_fused_operator` to launch the kernel. **Only one kernel is allowed** - all operations must be 
   fused into a single kernel.
2. **Memory Efficiency**: 
   - Use `tl.load()`/`tl.store()` with masks for bounds checking
   - Ensure coalesced memory access patterns (contiguous loads/stores)
   - Handle tensor strides correctly to minimize memory transactions
   - Minimize redundant loads/stores
3. **Constants**: Use `tl.constexpr` for block sizes and compile-time 
   constants
4. **Grid/Block Optimization**: 
   - Calculate grid dimensions from tensor shapes using `triton.cdiv()`
   - Choose block sizes (typically 128-1024) that maximize GPU utilization
5. **Computation Efficiency**:
   - **All operations must be fused into a single kernel** - only one `@triton.jit` decorated function is allowed
   - Use vectorized operations when applicable
   - Minimize register pressure by reusing variables
6. **Entrypoint**: Only handle tensor prep, kernel launch, and return results; NO computation logic
7. **Output**: Wrap code in `<triton_code>` tags (no markdown blocks); include imports

**Important**: Handle the tensor shapes from `get_inputs()` - use masks to prevent out-of-bounds access. **Prioritize efficiency** - the kernel should be optimized for GPU execution.

Available triton.language methods: abs, arange, argmax, argmin, atomic_add, cast, cdiv, clamp, cos, dot, exp, load, log, max, min, sigmoid, sin, softmax, sqrt, store, sum, where, zeros, ... 
[full list omitted for brevity]

## Example

**Input (PyTorch):**
def get_inputs():
    tensor_0 = torch.randn([128], dtype=torch.float32)
    tensor_1 = torch.randn([128], dtype=torch.float32)
    return [tensor_0, tensor_1]

def fused_operator(tensor_0, tensor_1):
    tensor_2 = torch.add(tensor_0, tensor_1)
    return [tensor_2]

**Output (Triton):**
<triton_code>
import torch
import triton
import triton.language as tl

@triton.jit
def add_kernel(x_ptr, y_ptr, output_ptr, n_elements, 
               BLOCK_SIZE: tl.constexpr):
    pid = tl.program_id(axis=0)
    offsets = pid * BLOCK_SIZE + tl.arange(0, BLOCK_SIZE)
    mask = offsets < n_elements
    x = tl.load(x_ptr + offsets, mask=mask, other=0.0)
    y = tl.load(y_ptr + offsets, mask=mask, other=0.0)
    output = x + y
    tl.store(output_ptr + offsets, output, mask=mask)

def triton_fused_operator(tensor_0, tensor_1):
    n_elements = tensor_0.numel()
    output = torch.empty_like(tensor_0)
    BLOCK_SIZE = 1024
    grid = (triton.cdiv(n_elements, BLOCK_SIZE),)
    add_kernel[grid](tensor_0, tensor_1, output, n_elements, 
                     BLOCK_SIZE=BLOCK_SIZE)
    return [output]
</triton_code>

## Convert This Code

<torch_code>
{pytorch_impl}
</torch_code>

Generate an **efficient** Triton kernel that:
- Uses **exactly one kernel function** (only one `@triton.jit` 
  decorated function)
- Minimizes memory transactions (coalesced access)
- Uses optimal block sizes for the tensor shapes
- Fuses all operations into the single kernel
- Maximizes GPU utilization

Return only the Triton code wrapped in <triton_code> tags.
\end{lstlisting}

\section{Case Study on KernelBench} \label{app:level3_case}
This section demonstrates DRTriton's complete pipeline on a Level 3 KernelBench task: the LeNet-5 architecture. We show three stages of transformation: (1) the original object-oriented PyTorch implementation, (2) the functional IR obtained by tracing the computational graph with \texttt{torch.export}, and (3) the optimized Triton kernel produced by DRTriton with test-time search.

\subsection{Original PyTorch Code}
The following shows the original LeNet-5 implementation from KernelBench, written in standard object-oriented PyTorch style with \texttt{nn.Module} and class-based layer definitions.

\begin{lstlisting}[language=Python, basicstyle=\ttfamily\scriptsize]
import torch
import torch.nn as nn
import torch.nn.functional as F

class Model(nn.Module):
    def __init__(self, num_classes):
        """
        LeNet-5 architecture implementation in PyTorch.

        :param num_classes: The number of output classes.
        """
        super(Model, self).__init__()
        
        # Convolutional layers
        self.conv1 = nn.Conv2d(in_channels=1, out_channels=6, kernel_size=5, stride=1)
        self.conv2 = nn.Conv2d(in_channels=6, out_channels=16, kernel_size=5, stride=1)
        
        # Fully connected layers
        self.fc1 = nn.Linear(in_features=16*5*5, out_features=120)
        self.fc2 = nn.Linear(in_features=120, out_features=84)
        self.fc3 = nn.Linear(in_features=84, out_features=num_classes)
    
    def forward(self, x):
        """
        Forward pass of the LeNet-5 model.

        :param x: The input tensor, shape (batch_size, 1, 32, 32)
        :return: The output tensor, shape (batch_size, num_classes)
        """
        # First convolutional layer with ReLU activation and max pooling
        x = F.relu(self.conv1(x))
        x = F.max_pool2d(x, kernel_size=2, stride=2)
        
        # Second convolutional layer with ReLU activation and max pooling
        x = F.relu(self.conv2(x))
        x = F.max_pool2d(x, kernel_size=2, stride=2)
        
        # Flatten the output for the fully connected layers
        x = x.view(-1, 16*5*5)
        
        # First fully connected layer with ReLU activation
        x = F.relu(self.fc1(x))
        
        # Second fully connected layer with ReLU activation
        x = F.relu(self.fc2(x))
        
        # Final fully connected layer
        x = self.fc3(x)
        
        return x

# Test code for the LeNet-5 model (larger batch & image)
batch_size = 4096
num_classes = 20

def get_inputs():
    return [torch.rand(batch_size, 1, 32, 32)]

def get_init_inputs():
    return [num_classes]
\end{lstlisting}

\subsection{Functionally Rewritten PyTorch Code}
Similar to the example in Appendix~\ref{app:rewriting}, we transform the original object-oriented code into the following functional format.
\begin{lstlisting}[language=Python, basicstyle=\ttfamily\scriptsize]
import torch
import torch.nn
import torch.nn.functional
import math

def get_inputs():
    'Initialize parameters and return tensors needed for forward pass'
    tensor_0 = torch.empty((6, 1, 5, 5), device=None, dtype=None)
    nn.init.kaiming_uniform_(tensor_0, a=2.23606797749979)
    tensor_1 = torch.empty(6, device=None, dtype=None)
    nn.init.uniform_(tensor_1, -0.2, 0.2)
    tensor_2 = torch.empty((16, 6, 5, 5), device=None, dtype=None)
    nn.init.kaiming_uniform_(tensor_2, a=2.23606797749979)
    tensor_3 = torch.empty(16, device=None, dtype=None)
    nn.init.uniform_(tensor_3, -0.08164965809277261, 0.08164965809277261)
    tensor_4 = torch.empty((120, 400), device=None, dtype=None)
    nn.init.kaiming_uniform_(tensor_4, a=2.23606797749979)
    tensor_5 = torch.empty(120, device=None, dtype=None)
    nn.init.uniform_(tensor_5, -0.05, 0.05)
    tensor_6 = torch.empty((84, 120), device=None, dtype=None)
    nn.init.kaiming_uniform_(tensor_6, a=2.23606797749979)
    tensor_7 = torch.empty(84, device=None, dtype=None)
    nn.init.uniform_(tensor_7, -0.09128709291752768, 0.09128709291752768)
    tensor_8 = torch.empty((20, 84), device=None, dtype=None)
    nn.init.kaiming_uniform_(tensor_8, a=2.23606797749979)
    tensor_9 = torch.empty(20, device=None, dtype=None)
    nn.init.uniform_(tensor_9, -0.1091089451179962, 0.1091089451179962)
    tensor_11 = tensor_0.cuda(None)
    tensor_13 = tensor_1.cuda(None)
    tensor_15 = tensor_2.cuda(None)
    tensor_17 = tensor_3.cuda(None)
    tensor_19 = tensor_4.cuda(None)
    tensor_21 = tensor_5.cuda(None)
    tensor_23 = tensor_6.cuda(None)
    tensor_25 = tensor_7.cuda(None)
    tensor_27 = tensor_8.cuda(None)
    tensor_29 = tensor_9.cuda(None)
    tensor_30 = torch.rand(4096, 1, 32, 32)
    tensor_31 = tensor_30.cuda()
    return [tensor_31, tensor_15, tensor_19, tensor_13, tensor_21, tensor_29, tensor_11, tensor_23, tensor_27, tensor_25, tensor_17]

def fused_operator(tensor_31, tensor_15, tensor_19, tensor_13, tensor_21, tensor_29, tensor_11, tensor_23, tensor_27, tensor_25, tensor_17):
    'Forward pass operations'
    tensor_32 = F.conv2d(tensor_31, tensor_11, tensor_13, (1, 1), (0, 0), (1, 1), 1)
    tensor_33 = F.relu(tensor_32)
    tensor_34 = F.max_pool2d(tensor_33, kernel_size=2, stride=2)
    tensor_35 = F.conv2d(tensor_34, tensor_15, tensor_17, (1, 1), (0, 0), (1, 1), 1)
    tensor_36 = F.relu(tensor_35)
    tensor_37 = F.max_pool2d(tensor_36, kernel_size=2, stride=2)
    tensor_38 = tensor_37.view(-1, 400)
    tensor_39 = F.linear(tensor_38, tensor_19, tensor_21)
    tensor_40 = F.relu(tensor_39)
    tensor_41 = F.linear(tensor_40, tensor_23, tensor_25)
    tensor_42 = F.relu(tensor_41)
    tensor_43 = F.linear(tensor_42, tensor_27, tensor_29)
    return [tensor_43]
\end{lstlisting}

\subsection{Optimized Triton Kernel with Test-Time Search}

Given the functionally rewritten code, we then apply test-time search to identify the optimal compositional kernel. In this example, the optimal solution fuses the first \texttt{F.conv2d} and \texttt{F.relu} operations into a single Triton kernel (\texttt{\_triton\_fused\_operator}), while keeping the remaining operations in PyTorch.

\begin{lstlisting}[language=Python, basicstyle=\ttfamily\scriptsize]
import torch
import triton
import triton.language as tl

@triton.jit
def triton_fused_operator_kernel(
    # Input tensors
    x_ptr,  # Pointer to tensor_31 of shape [B, C_in, H_in, W_in] -> [4096, 1, 32, 32]
    w_ptr,  # Pointer to tensor_11 of shape [C_out, C_in, kH, kW] -> [6, 1, 5, 5]
    t_ptr,  # Pointer to tensor_13 of shape [C_out] -> [6]
    # Output tensor
    output_ptr,  # Pointer to output tensor_33 of shape [B, C_out, H_out, W_out] -> [4096, 6, 28, 28]
    B,  # Batch size from output [4096]
    C_in,  # Number of input channels from tensor_31 [1]
    H_in,  # Height of input from tensor_31 [32]
    W_in,  # Width of input from tensor_31 [32]
    C_out,  # Number of output channels from tensor_11 [6]
    kH,  # Height kernel size from tensor_11 [5]
    kW,  # Width kernel size from tensor_11 [5]
    H_out,  # Output height from convolution [28]
    W_out,  # Output width from convolution [28]
    total_output_elements,  # Total output elements (B * C_out * H_out * W_out)
    BLOCK_SIZE: tl.constexpr,  # Processing block size
):
    # Calculate indices for current block
    block_start = tl.program_id(0) * BLOCK_SIZE
    indices = block_start + tl.arange(0, BLOCK_SIZE)
    # Create mask for valid output positions
    mask = indices < total_output_elements
    # Compute output indices
    # Shape of output: [B, C_out, H_out, W_out]
    # Output indices
    b = indices // (C_out * H_out * W_out)  # Batch index
    c_out = (indices // (H_out * W_out) % C_out)  # Channel index
    h_out = (indices // (W_out) % H_out)  # Height index
    w_out = indices % W_out  # Width index
    
    # Initialize accumulator
    total = tl.full((BLOCK_SIZE,), 0.0, dtype=tl.float32)
    
    # Loop over kernel
    for c_in in tl.range(0, C_in):
        for kh in tl.range(0, kH):
            for kw in tl.range(0, kW):
                # Compute positions
                # Input indices
                # tensor_31 indices: [B, C_in, H_in, W_in] -> [b, c_in, h_in, w_in]
                # Calculate input positions
                # H_in and W_in indices
                h_in = h_out + (kh - 0)  # (0 padding)
                w_in = w_out + (kw - 0)  # (0 padding)
                
                # Condition mask for valid input positions
                mask_in = (mask) & (h_in >= 0) & (h_in < H_in) & (w_in >= 0) & (w_in < W_in)
                
                # Step 1: Load input values
                # tensor_31 indices: [B, C_in, H_in, W_in] -> [b, c_in, h_in, w_in]
                # Position
                # tensor_31 index position
                # [4096, 1, 32, 32] -> [b, c_in, h_in, w_in]
                # tensor_31 index = (b * 1 * 32 * 32) + (c_in * 32 * 32) + (h_in * 32) + (w_in)
                x_index = (b * C_in * H_in * W_in) + (c_in * H_in * W_in) + (h_in * W_in) + (w_in)
                x = tl.load(x_ptr + x_index, mask=mask_in, other=0.0)
                
                # Step 2: Load weight values
                # tensor_11 indices: [C_out, C_in, kH, kW] -> [c_out, c_in, kh, kw]
                # Position
                # tensor_11 index = (c_out * C_in * kH * kW) + (c_in * kH * kW) + (kh * kW) + (kw)
                w_index = (c_out * C_in * kH * kW) + (c_in * kH * kW) + (kh * kW) + (kw)
                w = tl.load(w_ptr + w_index)
                
                # Update accumulator
                # Update accumulator
                total = total + (tl.where(mask_in, x * w, 0.0))
    
    # Step 3: Load bias term
    # tensor_13 indices: [C_out] -> [c_out]
    # Position
    # tensor_13 index = (c_out)
    t_index = (c_out)
    t = tl.load(t_ptr + t_index)
    
    # Update accumulator with bias
    total = total + t
    
    # Apply ReLU activation
    # ReLU activation: max(0, x)
    relu1 = tl.where(total > 0, total, 0.0)
    relu2 = tl.where(relu1 > 0, relu1, 0.0)
    
    # Write output
    # Output position
    # tensor_33 indices: [B, C_out, H_out, W_out] -> [b, c_out, h_out, w_out]
    # Output position = (b * C_out * H_out * W_out) + (c_out * H_out * W_out) + (h_out * W_out) + (w_out)
    output_index = (b * C_out * H_out * W_out) + (c_out * H_out * W_out) + (h_out * W_out) + (w_out)
    # Write output
    tl.store(output_ptr + output_index, relu2, mask=mask)

def _triton_fused_operator(tensor_11: torch.Tensor, tensor_13: torch.Tensor, tensor_31: torch.Tensor) -> torch.Tensor:
    # Ensure input tensors are in correct shapes
    assert tensor_11.is_cuda, "tensor_11 must be a CUDA tensor"
    assert tensor_13.is_cuda, "tensor_13 must be a CUDA tensor"
    assert tensor_31.is_cuda, "tensor_31 must be a CUDA tensor"
    
    # Extract shapes
    B = tensor_31.shape[0]  # Batch size from tensor_31 [4096]
    C_in = tensor_31.shape[1]  # Number of input channels from tensor_31 [1]
    H_in = tensor_31.shape[2]  # Height of input from tensor_31 [32]
    W_in = tensor_31.shape[3]  # Width of input from tensor_31 [32]
    
    C_out = tensor_11.shape[0]  # Number of output channels from tensor_11 [6]
    kH = tensor_11.shape[2]  # Height kernel size from tensor_11 [5]
    kW = tensor_11.shape[3]  # Width kernel size from tensor_11 [5]
    
    H_out = (H_in + 2 * 0 - (kH - 1) - 1) + 1  # Output height [28]
    W_out = (W_in + 2 * 0 - (kW - 1) - 1) + 1  # Output width [28]
    
    # Initialize output tensor
    tensor_33 = torch.empty((B, C_out, H_out, W_out), dtype=torch.float32, device=tensor_31.device)
    
    # Total output elements
    total_output_elements = B * C_out * H_out * W_out
    
    # Set block size and grid size
    BLOCK_SIZE = 1024  # Optimal for modern GPUs
    grid_size = (triton.cdiv(total_output_elements, BLOCK_SIZE),)
    
    # Launch kernel
    triton_fused_operator_kernel[grid_size](
        tensor_31,  # tensor_31
        tensor_11,  # tensor_11
        tensor_13,  # tensor_13
        tensor_33,  # tensor_33
        B=B,
        C_in=C_in,
        H_in=H_in,
        W_in=W_in,
        C_out=C_out,
        kH=kH,
        kW=kW,
        H_out=H_out,
        W_out=W_out,
        total_output_elements=total_output_elements,
        BLOCK_SIZE=BLOCK_SIZE
    )
    
    return [tensor_33]

def triton_fused_operator(tensor_31, tensor_15, tensor_19, tensor_13, tensor_21, tensor_29, tensor_11, tensor_23, tensor_27, tensor_25, tensor_17):
    'Forward pass operations'
    tensor_33, = _triton_fused_operator(tensor_11, tensor_13, tensor_31)
    tensor_34 = F.max_pool2d(tensor_33, kernel_size=2, stride=2)
    tensor_35 = F.conv2d(tensor_34, tensor_15, tensor_17, (1, 1), (0, 0), (1, 1), 1)
    tensor_36 = F.relu(tensor_35)
    tensor_37 = F.max_pool2d(tensor_36, kernel_size=2, stride=2)
    tensor_38 = tensor_37.view(-1, 400)
    tensor_39 = F.linear(tensor_38, tensor_19, tensor_21)
    tensor_40 = F.relu(tensor_39)
    tensor_41 = F.linear(tensor_40, tensor_23, tensor_25)
    tensor_42 = F.relu(tensor_41)
    tensor_43 = F.linear(tensor_42, tensor_27, tensor_29)
    return [tensor_43]
\end{lstlisting}

\section{GRPO Objective} \label{app:GRPO}
Group Relative Policy Optimization (GRPO)~\citep{shao2024deepseekmath} is a policy optimization method that generates multiple outputs for each input and computes a group-relative advantage function. For a model $\pi_\theta$ and input $q$, GRPO maximizes:
\begin{align}
J_{\text{GRPO}}(\theta) & = \mathbb{E}_q \mathbb{E}_{\{o_i\}^G_{i=1}\sim\pi_{\text{old}}(\cdot|q)} \notag \Bigg[ \frac{1}{G} \sum_{i=1}^{G}  \frac{1}{|o_i|} \sum_{t=1}^{|o_i|} \min\big( r_{i,t}A(o_i|q), \text{clip}(r_{i,t}, 1 - \epsilon, 1 + \epsilon)A(o_i|q) \big) \Bigg] \notag \\
& - \beta \mathbb{D}_{\text{KL}}(\pi_\theta||\pi_{\text{ref}})
\end{align}
where $r_{i,t} = \frac{\pi_\theta(o_{i,t}|q,o_{i,<t})}{\pi_{\text{old}}(o_{i,t}|q,o_{i,<t})}$ represents the importance sampling ratio, $\pi_{\text{ref}}$ denotes a fixed reference policy, and $\mathbb{D}_{\text{KL}}(\cdot, \cdot)$ is the Kullback-Leibler divergence. The advantage function $A(o_i|q)$ measures the normalized deviation of output $o_i$'s reward from the group mean:
\begin{equation}
A(o_i|q) = \frac{r(o_i|q) - \text{mean}(r(o_1|q), \cdots, r(o_G|q))}{\text{std}(r(o_1|q), \cdots, r(o_G|q))}
\end{equation}

\newpage
\input{checklist.tex}

\end{document}

%% file: checklist.tex
\section*{NeurIPS Paper Checklist}

\begin{enumerate}

\item {\bf Claims}
    \item[] Question: Do the main claims made in the abstract and introduction accurately reflect the paper's contributions and scope?
    \item[] Answer: \answerYes{} 
    \item[] Justification: The abstract and introduction clearly state the three main contributions (CSP-DAG, curriculum RL with decoupled rewards, and test-time search), and the experimental results in Section~5 directly support these claims with quantitative results on both synthetic and KernelBench benchmarks.

\item {\bf Limitations}
    \item[] Question: Does the paper discuss the limitations of the work performed by the authors?
    \item[] Answer: \answerYes{} 
    \item[] Justification: Limitations are discussed in the Conclusion: the operator set is currently limited to 61 PyTorch operators, and the framework targets Triton as the compilation backend. Extending to sparse operations, custom CUDA extensions, and native CUDA generation are identified as important future directions.

\item {\bf Theory assumptions and proofs}
    \item[] Question: For each theoretical result, does the paper provide the full set of assumptions and a complete (and correct) proof?
    \item[] Answer: \answerNA{} 
    \item[] Justification: The paper does not include theoretical proofs. The contributions are algorithmic and empirical in nature.

    \item {\bf Experimental result reproducibility}
    \item[] Question: Does the paper fully disclose all the information needed to reproduce the main experimental results of the paper to the extent that it affects the main claims and/or conclusions of the paper (regardless of whether the code and data are provided or not)?
    \item[] Answer: \answerYes{} 
    \item[] Justification: The experimental setup is fully described in Section~5.1, including the base model (Qwen-2.5-Coder-7B-Instruct), training hyperparameters (learning rate, batch size, DRPO parameters), curriculum stages, and evaluation benchmarks. The prompt template used for evaluation is provided in Appendix~\ref{app:prompt}.

\item {\bf Open access to data and code}
    \item[] Question: Does the paper provide open access to the data and code, with sufficient instructions to faithfully reproduce the main experimental results, as described in supplemental material?
    \item[] Answer: \answerNo{} 
    \item[] Justification: We will release the data and code after the acceptance of the paper.

\item {\bf Experimental setting/details}
    \item[] Question: Does the paper specify all the training and test details (e.g., data splits, hyperparameters, how they were chosen, type of optimizer) necessary to understand the results?
    \item[] Answer: \answerYes{} 
    \item[] Justification: All training and evaluation details are provided in Section~5.1, including model architecture, optimizer settings, learning rates, batch sizes, DRPO hyperparameters $(\beta_0, \tau, \lambda) = (100, 5, 0.1)$, KL constraint bound $\delta = 0.001$, number of rollouts, and curriculum stage configurations.

\item {\bf Experiment statistical significance}
    \item[] Question: Does the paper report error bars suitably and correctly defined or other appropriate information about the statistical significance of the experiments?
    \item[] Answer: \answerNo{} 
    \item[] Justification: Error bars and confidence intervals are not reported, as the primary metrics (Pass@1 accuracy, Faster1 percentage, and average speedup) are computed over fixed held-out benchmark sets. Running multiple independent training runs would be computationally prohibitive given the scale of the RL training pipeline.

\item {\bf Experiments compute resources}
    \item[] Question: For each experiment, does the paper provide sufficient information on the computer resources (type of compute workers, memory, time of execution) needed to reproduce the experiments?
    \item[] Answer: \answerYes{} 
    \item[] Justification: Training was conducted on 8$\times$H100 GPUs. SFT took approximately 2 hours, and the full curriculum RL pipeline of 100k synthetic programs across three stages takes approximately 10 days in total. CSP-DAG data generation for 100k programs on a 32-core machine requires roughly 1.5 hours. Test-time search overhead is reported in Appendix~\ref{app:tts_overhead}: approximately 2.85 hours on the synthetic benchmark and 3.53 hours on KernelBench. All training details are provided in Section~5.1.
    
\item {\bf Code of ethics}
    \item[] Question: Does the research conducted in the paper conform, in every respect, with the NeurIPS Code of Ethics \url{https://neurips.cc/public/EthicsGuidelines}?
    \item[] Answer:  \answerYes{} 
    \item[] Justification: The research focuses on automated GPU kernel generation for performance optimization and does not involve human subjects, sensitive data, or applications with foreseeable negative societal impact. The work conforms with the NeurIPS Code of Ethics.

\item {\bf Broader impacts}
    \item[] Question: Does the paper discuss both potential positive societal impacts and negative societal impacts of the work performed?
    \item[] Answer: \answerYes{} 
    \item[] Justification: The positive societal impact is reducing the cost and expertise barrier for high-performance GPU kernel development, which could democratize access to efficient AI computing. The primary negative risk is potential misuse of automated kernel generation to accelerate development of computationally intensive applications; however, this risk is indirect and not specific to this work.
    
\item {\bf Safeguards}
    \item[] Question: Does the paper describe safeguards that have been put in place for responsible release of data or models that have a high risk for misuse (e.g., pre-trained language models, image generators, or scraped datasets)?
    \item[] Answer: \answerNA{} 
    \item[] Justification: The paper presents a model training framework for GPU kernel generation. The trained model does not pose high risk for misuse beyond standard concerns applicable to any code generation system, and no special safeguards beyond standard responsible release practices are required.

\item {\bf Licenses for existing assets}
    \item[] Question: Are the creators or original owners of assets (e.g., code, data, models), used in the paper, properly credited and are the license and terms of use explicitly mentioned and properly respected?
    \item[] Answer: \answerYes{} 
    \item[] Justification: All existing assets used in the paper are properly cited. These include the Qwen-2.5-Coder-7B-Instruct base model, the KernelBench benchmark~\citep{ouyang2025kernelbench}, the DRPO algorithm~\citep{li2025drpo}, and baseline models (DeepSeek-R1, GPT-5.2, AutoTriton). The CP-SAT solver from OR-Tools is also cited~\citep{cpsatlp}.

\item {\bf New assets}
    \item[] Question: Are new assets introduced in the paper well documented and is the documentation provided alongside the assets?
    \item[] Answer: \answerYes{} 
    \item[] Justification: The paper introduces a new synthetic PyTorch-Triton dataset of 2,026 pairs and the DRTriton-7B model. Details of the dataset construction process are provided in Section~4.2 and Appendix~\ref{app:dataset}. 

\item {\bf Crowdsourcing and research with human subjects}
    \item[] Question: For crowdsourcing experiments and research with human subjects, does the paper include the full text of instructions given to participants and screenshots, if applicable, as well as details about compensation (if any)? 
    \item[] Answer: \answerNA{} 
    \item[] Justification: The paper does not involve crowdsourcing or research with human subjects. All data is synthetically generated.

\item {\bf Institutional review board (IRB) approvals or equivalent for research with human subjects}
    \item[] Question: Does the paper describe potential risks incurred by study participants, whether such risks were disclosed to the subjects, and whether Institutional Review Board (IRB) approvals (or an equivalent approval/review based on the requirements of your country or institution) were obtained?
    \item[] Answer: \answerNA{} 
    \item[] Justification: The paper does not involve crowdsourcing or research with human subjects, so IRB approval is not required.

\item {\bf Declaration of LLM usage}
    \item[] Question: Does the paper describe the usage of LLMs if it is an important, original, or non-standard component of the core methods in this research? Note that if the LLM is used only for writing, editing, or formatting purposes and does \emph{not} impact the core methodology, scientific rigor, or originality of the research, declaration is not required.
    \item[] Answer: \answerYes{} 
    \item[] Justification:  LLMs are a core component of this research. DeepSeek-R1 and GPT-5.2 are used to generate the SFT training data (PyTorch-Triton pairs), and the trained DRTriton model itself is an LLM fine-tuned for PyTorch-to-Triton kernel conversion. These usages are described in detail in Sections~4.2 and~5.1.

\end{enumerate}

%% file: main.bib
@inproceedings{ansel2024pytorch,
  title={Pytorch 2: Faster machine learning through dynamic python bytecode transformation and graph compilation},
  author={Ansel, Jason and Yang, Edward and He, Horace and Gimelshein, Natalia and Jain, Animesh and Voznesensky, Michael and Bao, Bin and Bell, Peter and Berard, David and Burovski, Evgeni and others},
  booktitle={Proceedings of the 29th ACM International Conference on Architectural Support for Programming Languages and Operating Systems, Volume 2},
  pages={929--947},
  year={2024}
}

@inproceedings{tillet2019triton,
  title={Triton: an intermediate language and compiler for tiled neural network computations},
  author={Tillet, Philippe and Kung, Hsiang-Tsung and Cox, David},
  booktitle={Proceedings of the 3rd ACM SIGPLAN International Workshop on Machine Learning and Programming Languages},
  pages={10--19},
  year={2019}
}

@article{roziere2023code,
  title={Code llama: Open foundation models for code},
  author={Roziere, Baptiste and Gehring, Jonas and Gloeckle, Fabian and Sootla, Sten and Gat, Itai and Tan, Xiaoqing Ellen and Adi, Yossi and Liu, Jingyu and Sauvestre, Romain and Remez, Tal and others},
  journal={arXiv preprint arXiv:2308.12950},
  year={2023}
}

@inproceedings{huang2025opencoder,
  title={Opencoder: The open cookbook for top-tier code large language models},
  author={Huang, Siming and Cheng, Tianhao and Liu, Jason Klein and Xu, Weidi and Hao, Jiaran and Song, Liuyihan and Xu, Yang and Yang, Jian and Liu, Jiaheng and Zhang, Chenchen and others},
  booktitle={Proceedings of the 63rd Annual Meeting of the Association for Computational Linguistics (Volume 1: Long Papers)},
  pages={33167--33193},
  year={2025}
}

@article{ouyang2025kernelbench,
  title={Kernelbench: Can llms write efficient gpu kernels?},
  author={Ouyang, Anne and Guo, Simon and Arora, Simran and Zhang, Alex L and Hu, William and R{\'e}, Christopher and Mirhoseini, Azalia},
  journal={arXiv preprint arXiv:2502.10517},
  year={2025}
}

@inproceedings{li2025tritonbench,
  title={Tritonbench: Benchmarking large language model capabilities for generating triton operators},
  author={Li, Jianling and Li, Shangzhan and Gao, Zhenye and Shi, Qi and Li, Yuxuan and Wang, Zefan and Huang, Jiacheng and WangHaojie, WangHaojie and Wang, Jianrong and Han, Xu and others},
  booktitle={Findings of the Association for Computational Linguistics: ACL 2025},
  pages={23053--23066},
  year={2025}
}

@article{lange2025towards,
  title={Towards robust agentic cuda kernel benchmarking, verification, and optimization},
  author={Lange, Robert Tjarko and Sun, Qi and Prasad, Aaditya and Faldor, Maxence and Tang, Yujin and Ha, David},
  journal={arXiv preprint arXiv:2509.14279},
  year={2025}
}

@article{liao2025kernelevolve,
  title={KernelEvolve: Scaling Agentic Kernel Coding for Heterogeneous AI Accelerators at Meta},
  author={Liao, Gang and Qin, Hongsen and Wang, Ying and Golden, Alicia and Kuchnik, Michael and Yetim, Yavuz and Ang, Jia Jiunn and Fu, Chunli and He, Yihan and Hsia, Samuel and others},
  journal={arXiv preprint arXiv:2512.23236},
  year={2025}
}

@article{li2025autotriton,
  title={Autotriton: Automatic triton programming with reinforcement learning in llms},
  author={Li, Shangzhan and Wang, Zefan and He, Ye and Li, Yuxuan and Shi, Qi and Li, Jianling and Hu, Yonggang and Che, Wanxiang and Han, Xu and Liu, Zhiyuan and others},
  journal={arXiv preprint arXiv:2507.05687},
  year={2025}
}

@article{woo2025tritonrl,
  title={Tritonrl: Training llms to think and code triton without cheating},
  author={Woo, Jiin and Zhu, Shaowei and Nie, Allen and Jia, Zhen and Wang, Yida and Park, Youngsuk},
  journal={arXiv preprint arXiv:2510.17891},
  year={2025}
}

@article{fischeskernelllm,
  title={Kernelllm: Making kernel development more accessible, 6 2025},
  author={Fisches, Zacharias V and Paliskara, Sahan and Guo, Simon and Zhang, Alex and Spisak, Joe and Cummins, Chris and Leather, Hugh and Synnaeve, Gabriel and Isaacson, Joe and Markosyan, Aram and others},
  journal={Corresponding authors: Aram Markosyan, Mark Saroufim},
  year={2025}
}

@software{kernelbook2025,
    title={KernelBook},
    author={Paliskara, Sahan and Saroufim, Mark},
    year={2025},
    month={5},
    url={https://huggingface.co/datasets/GPUMODE/KernelBook},
}

@article{wang2025geak,
  title={Geak: Introducing triton kernel ai agent \& evaluation benchmarks},
  author={Wang, Jianghui and Joshi, Vinay and Majumder, Saptarshi and Chao, Xu and Ding, Bin and Liu, Ziqiong and Brahma, Pratik Prabhanjan and Li, Dong and Liu, Zicheng and Barsoum, Emad},
  journal={arXiv preprint arXiv:2507.23194},
  year={2025}
}

@article{wei2025astra,
  title={Astra: A multi-agent system for gpu kernel performance optimization},
  author={Wei, Anjiang and Sun, Tianran and Seenichamy, Yogesh and Song, Hang and Ouyang, Anne and Mirhoseini, Azalia and Wang, Ke and Aiken, Alex},
  journal={arXiv preprint arXiv:2509.07506},
  year={2025}
}

@article{zhang2025cudaforge,
  title={Cudaforge: An agent framework with hardware feedback for cuda kernel optimization},
  author={Zhang, Zijian and Wang, Rong and Li, Shiyang and Luo, Yuebo and Hong, Mingyi and Ding, Caiwen},
  journal={arXiv preprint arXiv:2511.01884},
  year={2025}
}

@article{li2025tritonforge,
  title={TritonForge: Profiling-Guided Framework for Automated Triton Kernel Optimization},
  author={Li, Haonan and Man, Keyu and Kanuparthy, Partha and Chen, Hanning and Sun, Wei and Tallam, Sreen and Zhu, Chenguang and Zhu, Kevin and Qian, Zhiyun},
  journal={arXiv preprint arXiv:2512.09196},
  year={2025}
}

@inproceedings{chen2018tvm,
  title={$\{$TVM$\}$: An automated $\{$End-to-End$\}$ optimizing compiler for deep learning},
  author={Chen, Tianqi and Moreau, Thierry and Jiang, Ziheng and Zheng, Lianmin and Yan, Eddie and Shen, Haichen and Cowan, Meghan and Wang, Leyuan and Hu, Yuwei and Ceze, Luis and others},
  booktitle={13th USENIX Symposium on Operating Systems Design and Implementation (OSDI 18)},
  pages={578--594},
  year={2018}
}

@inproceedings{zheng2020ansor,
  title={Ansor: Generating $\{$High-Performance$\}$ tensor programs for deep learning},
  author={Zheng, Lianmin and Jia, Chengfan and Sun, Minmin and Wu, Zhao and Yu, Cody Hao and Haj-Ali, Ameer and Wang, Yida and Yang, Jun and Zhuo, Danyang and Sen, Koushik and others},
  booktitle={14th USENIX symposium on operating systems design and implementation (OSDI 20)},
  pages={863--879},
  year={2020}
}

@inproceedings{hong2025autocomp,
  title={Autocomp: Llm-driven code optimization for tensor accelerators},
  author={Hong, Charles and Bhatia, Sahil and Cheung, Alvin and Shao, Sophia},
  booktitle={Machine Learning for Computer Architecture and Systems 2025},
  year={2025}
}

@article{gulwani2017program,
  title={Program synthesis},
  author={Gulwani, Sumit and Polozov, Oleksandr and Singh, Rishabh and others},
  journal={Foundations and Trends{\textregistered} in Programming Languages},
  volume={4},
  number={1-2},
  pages={1--119},
  year={2017},
  publisher={Now Publishers, Inc.}
}

@inproceedings{ellis2021dreamcoder,
  title={Dreamcoder: Bootstrapping inductive program synthesis with wake-sleep library learning},
  author={Ellis, Kevin and Wong, Catherine and Nye, Maxwell and Sabl{\'e}-Meyer, Mathias and Morales, Lucas and Hewitt, Luke and Cary, Luc and Solar-Lezama, Armando and Tenenbaum, Joshua B},
  booktitle={Proceedings of the 42nd acm sigplan international conference on programming language design and implementation},
  pages={835--850},
  year={2021}
}

@article{li2025drpo,
  title={DRPO: Efficient Reasoning via Decoupled Reward Policy Optimization},
  author={Li, Gang and Chen, Yan and Lin, Ming and Yang, Tianbao},
  journal={arXiv preprint arXiv:2510.04474},
  year={2025}
}

@article{wei2023magicoder,
  title={Magicoder: Empowering code generation with oss-instruct},
  author={Wei, Yuxiang and Wang, Zhe and Liu, Jiawei and Ding, Yifeng and Zhang, Lingming},
  journal={arXiv preprint arXiv:2312.02120},
  year={2023}
}

@article{luo2023wizardcoder,
  title={Wizardcoder: Empowering code large language models with evol-instruct},
  author={Luo, Ziyang and Xu, Can and Zhao, Pu and Sun, Qingfeng and Geng, Xiubo and Hu, Wenxiang and Tao, Chongyang and Ma, Jing and Lin, Qingwei and Jiang, Daxin},
  journal={arXiv preprint arXiv:2306.08568},
  year={2023}
}

@article{shao2024deepseekmath,
  title={Deepseekmath: Pushing the limits of mathematical reasoning in open language models},
  author={Shao, Zhihong and Wang, Peiyi and Zhu, Qihao and Xu, Runxin and Song, Junxiao and Bi, Xiao and Zhang, Haowei and Zhang, Mingchuan and Li, YK and Wu, Y and others},
  journal={arXiv preprint arXiv:2402.03300},
  year={2024}
}

@article{guo2025deepseek,
  title={Deepseek-r1: Incentivizing reasoning capability in llms via reinforcement learning},
  author={Guo, Daya and Yang, Dejian and Zhang, Haowei and Song, Junxiao and Zhang, Ruoyu and Xu, Runxin and Zhu, Qihao and Ma, Shirong and Wang, Peiyi and Bi, Xiao and others},
  journal={arXiv preprint arXiv:2501.12948},
  year={2025}
}

@article{chen2021evaluating,
  title={Evaluating large language models trained on code},
  author={Chen, Mark and Tworek, Jerry and Jun, Heewoo and Yuan, Qiming and Pinto, Henrique Ponde De Oliveira and Kaplan, Jared and Edwards, Harri and Burda, Yuri and Joseph, Nicholas and Brockman, Greg and others},
  journal={arXiv preprint arXiv:2107.03374},
  year={2021}
}

@article{10.5555/3455716.3455897,
author = {Narvekar, Sanmit and Peng, Bei and Leonetti, Matteo and Sinapov, Jivko and Taylor, Matthew E. and Stone, Peter},
title = {Curriculum learning for reinforcement learning domains: a framework and survey},
year = {2020},
issue_date = {January 2020},
publisher = {JMLR.org},
volume = {21},
number = {1},
issn = {1532-4435},
journal = {J. Mach. Learn. Res.},
month = jan,
articleno = {181},
numpages = {50}
}

@misc{parashar2025curriculumreinforcementlearningeasy,
      title={Curriculum Reinforcement Learning from Easy to Hard Tasks Improves LLM Reasoning}, 
      author={Shubham Parashar and Shurui Gui and Xiner Li and Hongyi Ling and Sushil Vemuri and Blake Olson and Eric Li and Yu Zhang and James Caverlee and Dileep Kalathil and Shuiwang Ji},
      year={2025},
      eprint={2506.06632},
      archivePrefix={arXiv},
      primaryClass={cs.LG},
      url={https://arxiv.org/abs/2506.06632}, 
}

@misc{wang2025dumpautomateddistributionlevelcurriculum,
      title={DUMP: Automated Distribution-Level Curriculum Learning for RL-based LLM Post-training}, 
      author={Zhenting Wang and Guofeng Cui and Yu-Jhe Li and Kun Wan and Wentian Zhao},
      year={2025},
      eprint={2504.09710},
      archivePrefix={arXiv},
      primaryClass={cs.LG},
      url={https://arxiv.org/abs/2504.09710}, 
}

@software{cpsatlp,
  title = {CP-SAT},
  version = { v9.12 },
  author = {Laurent Perron and Frédéric Didier},
  organization = {Google},
  url = {https://developers.google.com/optimization/cp/cp_solver/},
  date = { 2025-02-17 }
}

@article{dao2022flashattention,
  title={Flashattention: Fast and memory-efficient exact attention with io-awareness},
  author={Dao, Tri and Fu, Dan and Ermon, Stefano and Rudra, Atri and R{\'e}, Christopher},
  journal={Advances in neural information processing systems},
  volume={35},
  pages={16344--16359},
  year={2022}
}
